\documentclass[%
aip,
amsmath,
amssymb,
preprint,%
]{revtex4-2}

\usepackage{graphicx}
\usepackage{dcolumn}
\usepackage{bm}

\usepackage[utf8]{inputenc}
\usepackage[T1]{fontenc}
\usepackage{mathptmx}
\usepackage{etoolbox}

\usepackage{subcaption}

\usepackage{eurosym}

\usepackage{comment}
\usepackage[colorinlistoftodos]{todonotes}

\usepackage{color}

\makeatletter
\def\@email#1#2{%
 \endgroup
 \patchcmd{\titleblock@produce}
  {\frontmatter@RRAPformat}
  {\frontmatter@RRAPformat{\produce@RRAP{*#1\href{mailto:#2}{#2}}}\frontmatter@RRAPformat}
  {}{}
}%
\makeatother
\begin{document}


\title[Forecasting with an N-dimensional Langevin equation and a neural-ordinary differential equation]{Forecasting with an N-dimensional
Langevin equation and a neural-ordinary differential equation}
\author{Antonio Malpica-Morales}
\email{a.malpica-morales21@imperial.ac.uk}
\affiliation{
Department of Chemical Engineering, Imperial College, London SW7 2AZ, United Kingdom
}%
\author{Miguel A. Dur\'an-Olivencia}
\email{m.duran-olivencia@imperial.ac.uk}
\affiliation{
Department of Chemical Engineering, Imperial College, London SW7 2AZ, United Kingdom
}%
\affiliation{
Research, Vortico Tech, M\'alaga 29100, Spain
}%
\author{Serafim Kalliadasis}
\email{s.kalliadasis@imperial.ac.uk}
\affiliation{
Department of Chemical Engineering, Imperial College,  London SW7 2AZ, United Kingdom
}%

\date{\today}

\begin{abstract}
\textbf{ABSTRACT}
\\
\\
Accurate prediction of electricity day-ahead prices is essential in
competitive electricity markets.
Although stationary electricity-price forecasting techniques have received
considerable attention, research on non-stationary methods is comparatively
scarce, despite the common prevalence of non-stationary features in
electricity markets.
Specifically, existing non-stationary techniques will often aim to address
individual non-stationary features in isolation, leaving aside the
exploration of concurrent multiple non-stationary effects.
Our overarching objective here is the formulation of a framework to
systematically model and forecast non-stationary electricity-price time
series, encompassing the broader scope of non-stationary behavior.
For this purpose we develop a data-driven model that combines an N-dimensional
Langevin equation (LE) with a neural-ordinary differential equation
(NODE).
The LE captures fine-grained details of the electricity-price behavior in
stationary regimes but is inadequate for non-stationary conditions.
To overcome this inherent limitation, we adopt a NODE approach to learn,
and at the same time predict, the difference between the actual
electricity-price time series and the simulated price trajectories
generated by the LE.
By learning this difference, the NODE reconstructs the non-stationary
components of the time series that the LE is not able to capture.
We exemplify the effectiveness of our framework using the Spanish
 electricity day-ahead market as a
prototypical case study.
Our findings reveal that the NODE nicely complements the LE, providing a
comprehensive strategy to tackle both stationary and non-stationary
electricity-price behavior.
The framework's dependability and robustness is demonstrated through
different non-stationary scenarios by comparing it against a range of basic
na\"ive methods.
\end{abstract}

\maketitle

\begin{quotation}
In the realm of time-series analysis, the term ``stationary" refers to those
observable quantities that exhibit consistent and predictable behavior over
time, while the term ``non-stationary" encompasses a time regime governed by
time-varying features.
Electricity-price time series alternate between both stationary and
non-stationary regimes because of the complex nature of electricity markets;
this is especially so with the day-ahead market which accounts for the
electricity trading on the following day.
To address this challenge, we propose a comprehensive data-driven framework to model and forecast
electricity prices independently of the prevailing market regime.
Our framework combines a stochastic differential equation which approximates
the time evolution of the price in stationary conditions, and a neural-ordinary
differential equation responsible for the reconstruction of the
elusive non-stationary components of the price that the stochastic equation
cannot capture by itself.
We illustrate the dependability, strength, and robustness of our framework
using data from the Spanish electricity day-ahead market and through various
non-stationary settings.
Not only does the proposed framework advance existing techniques to tackle
non-stationary electricity-price time series, but also offers a flexible and
versatile methodology with potential applications to other scientific domains
characterized by the coexistence of both stationary and non-stationary
phenomena.
\end{quotation}

\section{Introduction}
\label{sec:introduction}

Time series are ubiquitous in science and engineering.
Most time-series recordings register macroscopic/state variables of complex
dynamical systems that are composed of many interacting elements at a
microscopic level.
Describing such systems from first principles requires a large number of
degrees of freedom accounting for the non-trivial interactions between their
constituent components.
Not surprisingly, this high-dimensional description poses significant
challenges in terms of both modeling and computational tractability.
A way forward is coarse graining: the objective is to obtain an effective
description that (judiciously) averages out the microscopic properties and
retains the main effects at the macroscopic level. This then turns the study
of complex systems to that of examining the time evolution of their
macroscopic/state or coarse-grained variables.
Time-series analysis of the latter then aims to facilitate the understanding
and inference of relationships between observed macroscopic effects and the
underlying microscopic governing dynamics.~\cite{Friedrich2011,
Kalliadasis2015, Anvari2016}

Among the various time-series properties, stationarity is critical for the
understanding of the equilibrium state (and its possible time evolution) 
of the underlying complex system characterized by the time series.
A stationary time series is one for which the probability density function
(PDF) governing the time-series statistics is invariant in
time.~\cite{Brockwell1991}
In contrast, non-stationary time series exhibit a time-dependent PDF.
This temporal variability can arise due to either an explicit time dependence
of the moments of the stochastic process governing the underlying complex
system dynamics,~\cite{Kirchgassner2013} or multi-regime changes where
distinct PDFs govern the system dynamics in different
regimes.~\cite{Hamilton1989, Battaglia2011}

While stationary modeling has been extensively studied generating a
well-established literature,~\cite{DeGooijer2006, Box2016} non-stationary
techniques have not received the same level of attention.~\cite{Cheng2015}
This imbalance is primarily due to the difficulties in unraveling the causes
of the non-stationary behavior, as it often intertwines with intricate
nonlinear dynamics occurring at the system's constituent/microscopic level.~\cite{Huang1998, Boccara2010}
The frequent course of action when dealing with time series is to
transform a non-stationary signal into a quasi-stationary
one.~\cite{Wu2007}
This approach attenuates the nonlinear features, simplifying the time-series analysis.
However, it comes at the expense of losing detailed system interactions that
are embedded within the inherent nonlinear features.

An area that accounts for good part of the literature within the field of
time-series analysis is electricity-prices modeling,~\cite{Nogales2002,
Gonzalez2005, Aggarwal2009, Weron2014, Nowotarski2018, Lago2021,
Deschatre2021} rather topical these days because of the global energy crisis
that began in the aftermath of the COVID-19 pandemic.
Electricity, in particular, is a special commodity with limited economic
feasibility for storage among the different energy
resources.~\cite{Bessembinder2002}
Indeed, electricity markets and the electricity power-system operation must
guarantee a continuous equilibrium between electricity generation and demand,
ensuring the security of supply, which is of paramount importance in
competitive economies.

In the electricity day-ahead market, a daily auction is held where
electricity generators and demand agents submit their electricity energy
offers for each time-block auctioned.
Typically, in European and American markets, one time-block allocates 1 hour of the following day.
Thus, every day, the outcome of the day-ahead market is a 24-dimensional
array of electricity prices for the 24 h of the following day.
The electricity day-ahead market is a complex system affected by a myriad
of factors, such as fuel prices, weather conditions, or market players'
bidding strategies.
The nonlinear interactions between these, often competing, factors ultimately
translate into non-stationary electricity-price time series.
By addressing this non-stationary behavior, market players and policymakers
would be better equipped to understanding the price dynamics, and, as a
consequence, to improving the accuracy of their electricity-prices
forecasts.~\cite{Knittel2005}
This, in turn, would enhance their decision-making capabilities within the
competitive electricity market landscape.

Unraveling the dynamics of electricity markets requires a comprehensive
analysis of electricity-price time series encompassing their non-stationary
characteristics. In recent decades, there have been notable research efforts
aiming to address the challenges posed by non-stationary conditions in
electricity-price time series.~\cite{Weron2004, Borovkova2017, Qiu2017,
Kostrzewski2019, Mestre2021}
However, most previous work has focused only on individual non-stationary
features, such as price jumps or moment variations, without tackling multiple
non-stationary effects simultaneously.
Alternatively, as mentioned earlier, some studies transform the
non-stationary time series into a quasi-stationary signal by studying the
time evolution of the changes over consecutive time steps, thus eliminating
possible trend components.
Hence, non-stationary behavior remains elusive, and there still remains a
need for a methodology to account for it.
Our overarching objective here is precisely to develop a comprehensive
framework for the rational and systematic analysis of non-stationary
electricity-price time series regardless of the specific non-stationary
features exhibited.

The backbone of our approach lies in the decomposition of the electricity
day-ahead price time series into both stationary and non-stationary
signals.
For the stationary component, we adopt a data-driven model underpinned by
an N-dimensional Langevin equation (LE).~\cite{Zwanzig1973, Russo2022}
As we shall illustrate, this model captures fine details governing the price
dynamics, yielding a reliable approximation of the price evolution in
stationary conditions.
Consequently, any disparities observed between the price evolution
simulated by the LE and the actual electricity-price time series are
entirely driven by the non-stationary aspects of the electricity day-ahead market.
We extend the LE-based price model to comprehensively address these
non-stationary features by complementing it with a neural-ordinary
differential equation (NODE).~\cite{Chen2018, Kidger2021}
This extension involves training and validating the NODE to effectively
approximate the time evolution of the difference between the actual prices
and those predicted by the LE.
We then incorporate the results obtained from the NODE to the LE output to
forecast the electricity price.
This offers a novel and unique architecture for dealing with non-stationary
electricity day-ahead prices.
The conjoining of the LE and the NODE guarantees a trade-off between
interpretability and accurate forecast.
On the one hand, the LE enables an explicit price-equation formulation which unravels
the underlying price dynamics in stationary conditions.
On the other hand, the NODE exclusively predicts the price-signal term that
the LE cannot capture, shedding light on the non-stationary behavioral
patterns of the electricity day-ahead price.

\begin{figure*}
\captionsetup{justification=raggedright}
{\includegraphics[width=\textwidth]{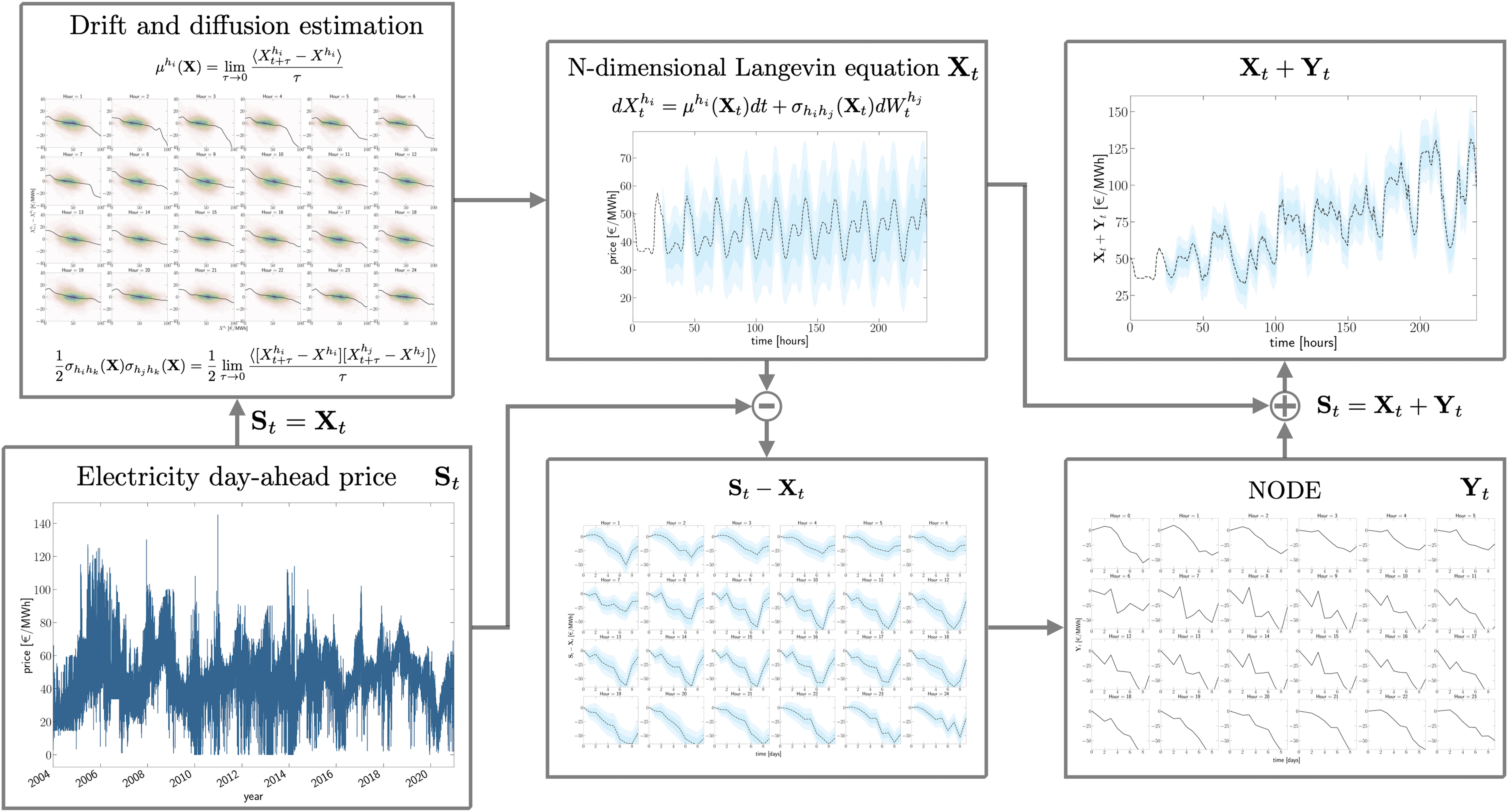}}
\caption{Schematic diagram of the framework proposed to forecast time series,
$\mathbf{S}_t$.
The framework consists of a two-stage process.
First, a N-dimensional LE approximates the stationary
component, $\mathbf{X}_t$, of $\mathbf{S}_t$.
Second, the NODE extends $\mathbf{X}_t$ to account for
the non-stationary behavior, $\mathbf{Y}_t$, of $\mathbf{S}_t$ that the LE cannot capture. The framework
is applied to the electricity day-ahead market, in which case $\mathbf{S}_t$ represents
the electricity day-ahead prices.
}
\label{fig:graph-framework}
\end{figure*}

Section~\ref{sec:methodology} introduces the methodology of the
proposed framework for modeling and predicting non-stationary
electricity day-ahead prices.
In Sec.~\ref{sec:case-study}, we exemplify the proposed framework with
the Spanish electricity day-ahead market.
By comparing the results of the LE model with the actual electricity-price
time series, we highlight the necessity to address the non-stationary
behavior.
We then demonstrate the NODE's effectiveness and efficiency to accommodating
the non-stationary effects and enhancing the electricity-price prediction
generated by the LE.
Finally, we contrast the performance of our framework in forecasting
non-stationary prices against a range of na\"ive techniques that rely on
heuristics.
Through this comparison we validate the reliability, robustness, and strength
of our framework.
Finally, Sec.~\ref{sec:conclusions} offers concluding remarks and suggestions for promising
future research directions.

\section{Methodology}
\label{sec:methodology}

Decomposing a time series into stationary and non-stationary
signals is a well-established procedure in the field of nonlinear
time-series analysis,~\cite{Huang1998, Verdes2001, Szeliga2003, Szeliga2003b,
Verdes2006, Hara2012} with applications in diverse fields ranging from
neuroscience \cite{vonBunau2010} to climate science.~\cite{Verdes2007}
Following this strategy, we propose the following decomposition of the
electricity price:
\begin{equation}
\mathbf{S}_{t} = \mathbf{X}_t + \mathbf{Y}_t,  \label{eq:price-representation}
\end{equation}
where $\mathbf{S}_t = \{ S_{t}^{h_i} \}$ is the collection of electricity
prices at hours $h_i=\{ 1, \ldots, 24\}$ of the day $t \in \mathbb{N}_{0}$,
$\mathbf{X}_t = \{ X_t^{h_i} \}$ is a multivariate stochastic process
accounting for the stationary part of the price evolution, and $\mathbf{Y}_t
= \{ Y_t^{h_i} \}$ is a deterministic time-dependent function embodying the
non-stationary effects.

Figure~\ref{fig:graph-framework} depicts a diagrammatic representation of our
framework.
As discussed in Sec.~\ref{sec:introduction},
the hourly electricity day-ahead price is the result of the daily auction process.
This process yields a hourly price which evolves in time (per day) exhibiting
an average trend and state-dependent fluctuations, under the assumption of a
moderate daily change of the demand and the available electricity generation.
To model this price evolution, we adopt a multivariate LE, a versatile prototypical model for the evolution of stochastic
variables in complex systems, including financial instruments and global
weather effects.~\cite{Russo2022}
This simple, yet effective, model approximates the stationary component of
the time evolution of the price, capturing the subtle price movements
observed across consecutive days.
However, it is not capable to account for substantial price variations arising
from complex market dynamics, such as persistent trends and changes in volatility.

To address this limitation, we adopt a statistical-learning technique, which we referred to in 
Sec.~\ref{sec:introduction} as NODE, to encompass the non-stationary behavior that the 
LE cannot capture.
Although classic neural networks (NNs) can be
easily customized and have demonstrated good performance in numerous settings
especially where the inputs and outputs are high-dimensional, including
time-series prediction,~\cite{Lim2021, Torres2021} they do generate a
discrete time-fixed prediction whose time scope depends on the design of the
network architecture.
Conversely, NODEs are increasingly gaining traction as an attractive
alternative machine-learning tool to model complex dynamics, with the
capability of treating time as a continuous variable regardless of the
network layout.
Specifically, NODEs aim at learning the underlying vector field dictating the
time evolution of a system at hand.~\cite{Kidger2021}
This means that the NODE approximates the unknown function governing the
time-dependent differential equation of a system's observables.
This concept seamlessly aligns with our overarching objective of unraveling
the dynamics of the price evolution.
As we will demonstrate, the price predictions obtained from the NODE enhance
substantially the forecasting capacity of the simulated price trajectories of
the LE.

It should be emphasized that the actual dynamic laws governing the day-ahead
price movement are not known exactly and might even be intractable.
Nevertheless, it is reasonable to assume that the electricity price follows a
predefined yet unknown dynamics at least for a short-term time scale, usually
spanning a few days.
These unknown dynamics can be partially uncovered by approximating the
time evolution of the electricity price by means 
of data-driven techniques.
This is precisely the core of our framework's rationale.
Our hypothesis is that the synergistic combination of the LE, supporting the
stationary features, $\mathbf{X}_t$, and the NODE, driving the non-stationary
behavior of the price signal, $\mathbf{Y}_t$, is sufficient to estimate the
prevailing features of the price evolution, $\mathbf{S}_t$.
Also, by separating $\mathbf{S}_t$ into $\mathbf{X}_t$ and $\mathbf{Y}_t$,
our framework facilitates the understanding of the underlying market dynamics
generating the time evolution of $\mathbf{S}_t$ in a short-term horizon.

\subsection{Stationary component: LE}
\label{subsec:gle}

We reconstruct $\mathbf{X}_t$ using a multivariate LE:
\begin{equation}
dX_t^{h_i} = \mu^{h_i}(\mathbf{X}_t)dt + \sigma_{h_i h_j}(\mathbf{X}_t) dW_t^{h_j}, \label{eq:gle}
\end{equation}
where $\mu^{h_i}$ and $\sigma_{h_i h_j}$ are the drift and diffusion
coefficients, respectively, with $h_j = \{ 1, \ldots, 24 \}$, and $W_t^{h_j}$
is a Wiener process vector with Gaussian increments: $W^{h_j}_{t+dt} -
W^{h_j}_{t} \sim \mathcal{N}(0, dt)$.
We estimate the drift and diffusion coefficients from empirical
price observations following the definition of the Kramers-Moyal expansion
coefficients.~\cite{Risken1966}
(See Appendix~\ref{app:drift-diffusion-estimation} for further details.)

The LE in Eq.~\eqref{eq:gle} is a valid approximation for $\mathbf{S}_t$, i.e., $\mathbf{S}_t = \mathbf{X}_t$,
only when the electricity day-ahead market is at equilibrium.
At this equilibrium state, we can assume that historical electricity
day-ahead prices are representative of the future price evolution, which is consistent
with the time-invariant PDF inherent in stationarity.
However, there are a myriad of both internal and
external factors that disrupt the normal operation of the electricity
day-ahead market, leaving the market in a non-equilibrium state, with prices
deviating from historical expected records.
Furthermore, the stationary assumption underlying Eq.~\eqref{eq:gle} is tightly
linked to the Markov property.
Under this property, simulations of the time evolution of the price depend
entirely on the present price level (the initial condition) and not on past
records.
Consequently, as it stands, the LE is impeding our ability to model and
forecast electricity prices accurately because it hinges solely on historical price values, it does not incorporate endogenous/exogenous factors that modify market behavior and operates as a memoryless process due to the Markov property.

\subsection{Non-stationary component: NODE}
\label{subsec:neural-differential-equation}

To alleviate the limitations of the LE and enhance its price approximation,
we complement it with a NODE to reconstruct $\mathbf{Y}_t$, as
represented in Eq.~\eqref{eq:price-representation}.
Within the neural differential equation family, NODEs~\cite{Chen2018} have
demonstrated remarkable performance as a general-purpose machine-learning
methodology for solving initial-value problems (IVPs) in deterministic
systems.~\cite{Lai2021, Kim2021, Chen2022, Dhadphale2022} NODEs utilize a
system of ODEs that generate trajectories of the macroscopic observables of
the system at hand.
Unlike classic fixed-time NNs, NODEs process the input data, i.e., the time
evolution of the observables, in a continuous manner, which in turn endows
NODEs with increased representation capabilities of complex temporal
patterns.
The basic idea is to parameterize the derivatives of the ODEs to be
solved using NNs, the output of which is evaluated using a (black-box) ODE
solver.
Adjusting the weights of the NNs through common gradient-based optimization
techniques, appropriate approximations of the unknown functions dictating the
system's dynamics are obtained.
This then enables NODEs to reconstruct the trajectories of the input data.
As the ODE solver can be arbitrarily set to any specific time horizon, the
NODE has the ability to predict up to any desired time, resulting in a
continuous-time generative model.

Referring back to Eq.~\eqref{eq:price-representation},
the signal $\mathbf{Y}_t = \mathbf{S}_t - \mathbf{X}_t$
contains the residual component of the price behavior not captured by
the stochastic process $\mathbf{X}_t$.
For a short-term period, we can approximate $\mathbf{Y}_t$ with a flexible
functional form. The combination of the NODEs, being a continuous-time
generative model, and the universal approximation capabilities of NNs, yields
a model for the continuous-time approximation of  $\mathbf{Y}_t$,
\begin{equation}
d\mathbf{Y}_t = f\left( \mathbf{Y}_t, t, \theta \right)dt, \quad \mathbf{Y}_{0} = \mathbf{Y}_{t_0}, \label{eq:nural-ode}
\end{equation}
where $f$ is a deterministic function (the vector field that governs the ODE
dynamics), $\theta$ are the weights of the proposed NN architecture, and
$\mathbf{Y}_{t_0}$ corresponds to the initial condition.
During the NN training stage, the weights, $\theta$, are updated iteratively
to approximate the true but unknown dynamics $f$ of the system.

Such an update process depends on the error metric, the so-called loss function.
The loss function, $\mathcal{L}$, measures the difference between the real
trajectory from $t_0$ to $t_1$, $\{\mathbf{Y}_{t_0}, \, \ldots, \,
\mathbf{Y}_{t_1}\}$, and the trajectory generated by the NN as it was the
solution of an IVP:
\begin{equation}
\mathcal{L}\left(\{\mathbf{Y}_{t_0}, \, \ldots, \, \mathbf{Y}_{t_1}\}, \,  \mathbf{Y}_{t_0} + \int_{t_0}^{t_1}  f\left( \mathbf{Y}_t, t, \theta \right) dt \right).
\label{eq:nn-loss}
\end{equation}

The calculation of $\mathbf{Y}_t$ and implementation of the NODE is as
follows:
considering an initial condition, $\mathbf{S}_{t_0}$,
we generate a set of $n$ independent random paths
for a predefined period of time $p$ solving numerically the LE as in Eq.~\eqref{eq:gle}.
These random paths produce a collection $\{ \mathbf{X}_{t_0}, \,
\mathbf{X}_{t_0+1}, \, \ldots , \, \mathbf{X}_{t_0+p} \}_{n}$ that we compare
with the actual electricity price $\{\mathbf{S}_{t_0}, \, \mathbf{S}_{t_0+1},
\, \ldots, \, \mathbf{S}_{t_0+p}  \}$ obtaining a dataset,
$\mathcal{D}^{p}_{n} = \{ \mathbf{S}_{t_0} -  \mathbf{X}_{t_0}, \,
\mathbf{S}_{t_0+1} - \mathbf{X}_{t_0+1}, \, \ldots, \, \mathbf{S}_{t_0+p} -
\mathbf{X}_{t_0+p} \}_{n}$.
We then use $\mathcal{D}^{p}_{n}$ to train the NODE.
Throughout the training process, the NODE learns $\mathbf{Y}_t$, attempting to
compensate the shortcomings of the LE in approximating $\mathbf{S}_t$.
Once the NODE learns the function $\mathbf{Y}_t$ over the period $p$, we
employ its continuous-time generative capability to predict the out-of-sample
values $\mathbf{Y}_{t}$ until time step $p+q, \, q > 0$.
Combining the out-of-sample predictions with the random paths of the LE from
Eq.~\eqref{eq:gle}, we can define the following test dataset $\mathcal{D}^{p,
q}_{n} = \{ \mathbf{X}_{t_0+p+1} + \mathbf{Y}_{t_0+p+1}, \, \ldots,
\mathbf{X}_{t_0+p+q} + \mathbf{Y}_{t_0+p+q} \}_{n}$.
By contrasting $\mathcal{D}^{p,q}_{n}$ with $\mathbf{S}_t$, we can validate
the performance of our framework in reconstructing and forecasting the
electricity day-ahead prices.

It is worth mentioning that the NODE is applied after obtaining the results of the
LE.
While implementing the NODE directly to model and forecast $\mathbf{S}_t$ may seem appealing, it poses significant challenges in terms of computational requirements,
explainability, and performance.
First, the NN architecture within the NODE must be sufficiently sophisticated (deep,
convolutional, or residual) to capture the information already unraveled by the LE
and simultaneously address time-dependent effects.
Another limitation of NODEs is their short-time scope.
Hence, to process the entire historical time series of $\mathbf{S}_t$,
the NODE should be trained and validated with a sliding window multiple times,
increasing computational demands and training times.
Moreover, since it operates as a black-box model, we could not extract and
understand the prevailing dynamics that dictate the time evolution of
$\mathbf{S}_t$, which the LE enables.
Finally, the NODE may suffer from overfitting, as the historical time series of $\mathbf{S}_t$ comprises a single sample available for the training procedure.
Conversely, the NODE within our framework uses the $n$ residual trajectories of
$\mathbf{S}_t - \mathbf{X}_t$, preventing from overfitting and ensuring good generalization.

\subsection{Na\"ive methods}
\label{subsec:naive-methods}
We also evaluate the effectiveness of our framework in comparison to
several na\"ive methods.
This benchmark, the so-called na\"ive test, is a widely recognized procedure to assess the
accuracy of the proposed electricity-price forecasting technique against
basic rule-based predictions.~\cite{Nogales2002, Weron2014, Cruz2011}
In our study, the na\"ive methods approximate $\mathbf{Y}_{t}$ using heuristic
formulations.
The purpose of the benchmark is to demonstrate the advantage of our framework
over these simplified approaches, verifying it as a proof of concept.
The first na\"ive approach, denoted as ``LE + 1 day difference", consists of
the following steps:
\begin{eqnarray}
\mathbf{Y}_{t_0+p+1} &&=\mathbf{S}_{t_0+p} - \mathbf{S}_{t_0+p-1} \nonumber \\
\mathbf{\hat{S}}_{t_0+p+1} &&= \mathbf{X}_{t_0+p+1} + \mathbf{Y}_{t_0+p+1}, \label{eq:first-naive}
\end{eqnarray}
while the second na\"ive approach, referred to as ``LE + initial condition
difference," is,
\begin{eqnarray}
\mathbf{Y}_{t_0+p+1} &&= \mathbf{S}_{t_0+p} - \mathbf{S}_{t_0} \nonumber \\
\mathbf{\hat{S}}_{t_0+p+1} &&= \mathbf{X}_{t_0+p+1} + \mathbf{Y}_{t_0+p+1}.
\end{eqnarray}

\begin{figure*}
\captionsetup{justification=raggedright}
\includegraphics[width=\textwidth]{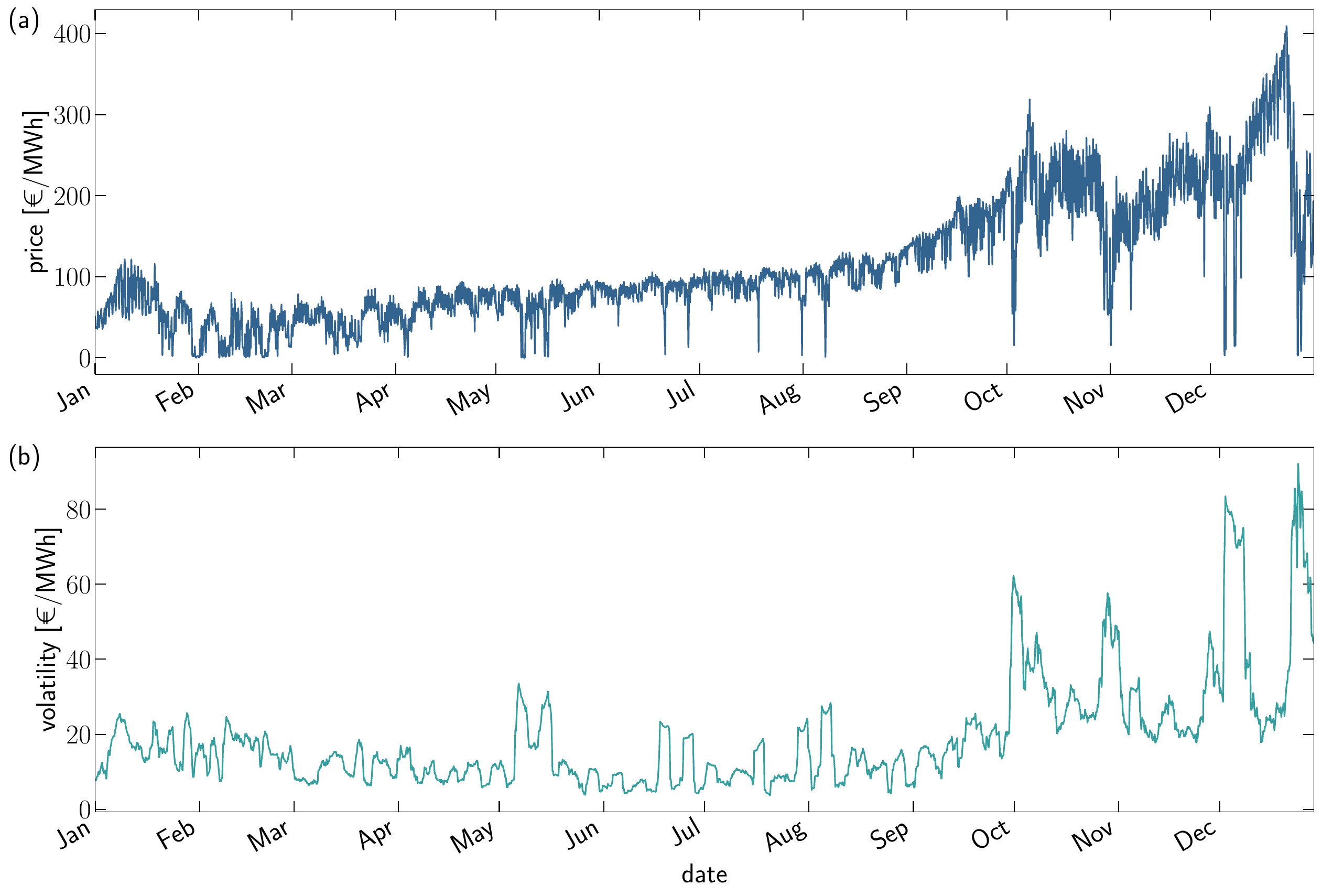}
\caption{(a) Spanish electricity day-ahead price time series and (b) its associated volatility for the year 2021.
The volatility is computed as the standard deviation of the last 3 days.}
\label{fig:2021-time-series}
\end{figure*}

\section{Case Study}
\label{sec:case-study}

We exemplify our framework using data from the Spanish electricity day-ahead
market for the period spanning 2004 - 2021.
The period from 2004 to 2020 constitutes the training dataset to estimate the
drift and diffusion coefficients of the LE in Eq.~\eqref{eq:gle}.
The time series from the training dataset
consists of the hourly electricity prices in
$\text{\euro}/\text{MWh}$, comprising a total of 149040 data
points with 6210 samples per hour.
We validate the electricity-price trajectories generated by the LE using the
Spanish electricity data for the year 2021.
This validation dataset corresponds to 8760 samples with 365 samples per hour.
Such validation motivates the extension of the LE with the NODE enhancing thus the
overall predictive capabilities of our framework.

\subsection{LE validation}
\label{subsec:gle-validation}

Figure~\ref{fig:2021-time-series} displays the Spanish electricity-price
time series and its associated volatility for the whole year 2021.
There are two noteworthy non-stationary effects.
First, trends that are maintained at irregular time intervals.
For example, a slight positive trend occurs at the beginning of the year
that subsequently disappears, leading to a price plunge around February.
There is also a persistent trend that starts around June and steadily grows
until about mid-October.
Second, changes in volatility levels, as shown in the bottom subplot where
volatility peaks arise in mid-May, the beginning of October, and throughout
December.

We now assess the LE performance in approximating the time evolution
of the Spanish electricity prices in 2021.
For this purpose, we simulate a collection of $n=10^3$ independent sample
paths of the LE in Eq.~\eqref{eq:gle} using the Euler-Maruyama
method~\cite{kloeden_1992} with a time interval of $ T = p = 9$ days and a
step size of 1 day.
By applying the Euler-Maruyama method on different days, i.e., different
initial conditions, we can analyze the LE behavior and compare it to the
time evolution of the real electricity price for specific scenarios.

\begin{figure}[t]
\centering
\captionsetup{justification=raggedright}
\includegraphics[width=0.7\textwidth]{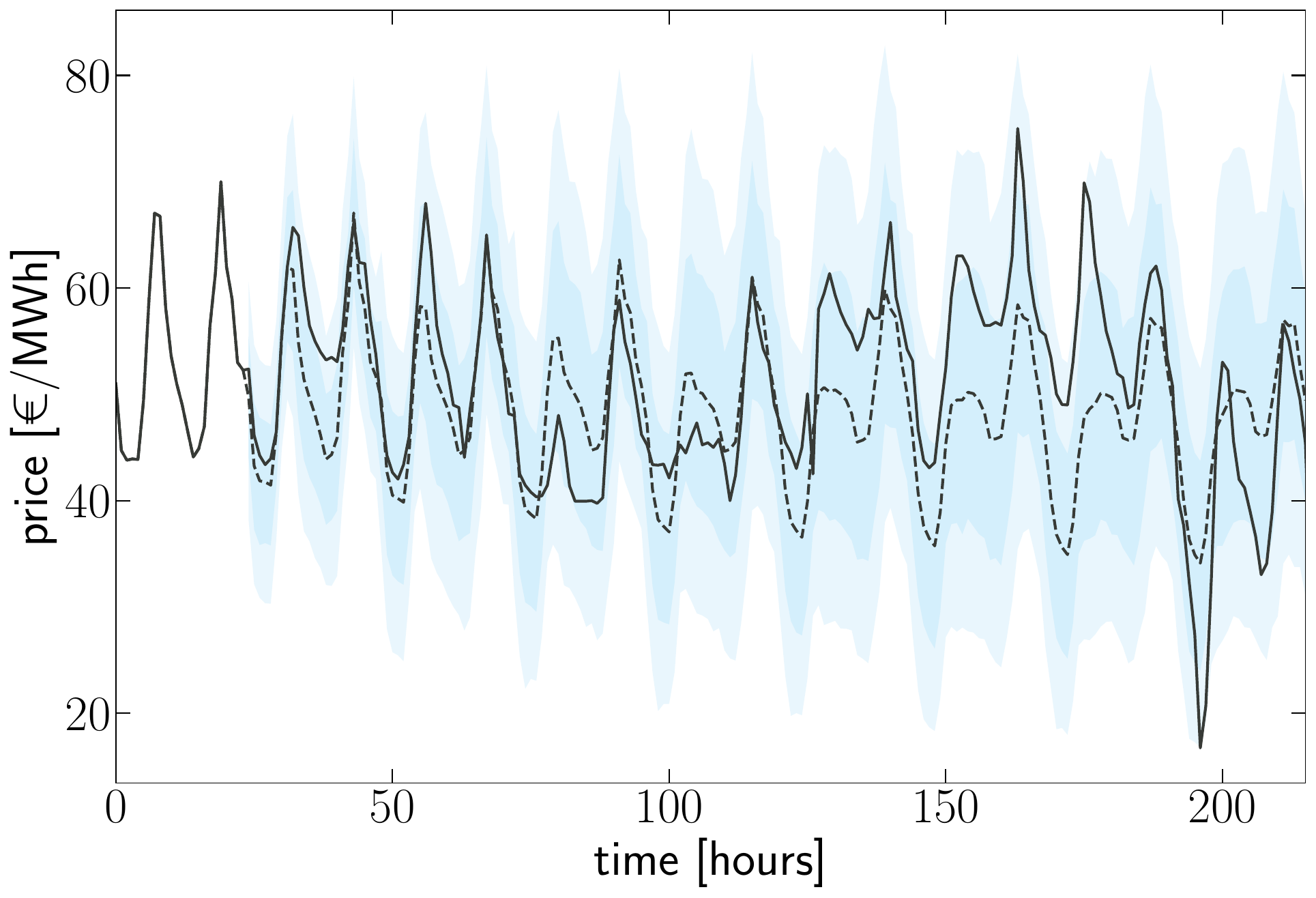}
\caption{Simulation of $\mathbf{X}_t$ obtained from the LE in
stationary conditions.
Initial date: 3/3/2021.
Solid line is the true electricity price, $\mathbf{S}_t$.
Dashed line corresponds to the mean price over $10^{3}$ simulated paths of $\mathbf{X}_t$.
Shaded areas delimit the following percentile ranges: [25, 75] (dark)
and [10, 90] (light).
Thus, the lightest area at the bottom of the plot encloses the percentiles
[10, 25], while the lightest area at the top encloses the percentiles
[75, 90].
}
\label{fig:stationary-gle}
\end{figure}

\begin{figure*}[t]
\centering
\captionsetup{justification=raggedright}

\mbox{}\hfill
\begin{subfigure}[t]{0.35\textwidth}
\includegraphics[width=1\textwidth]{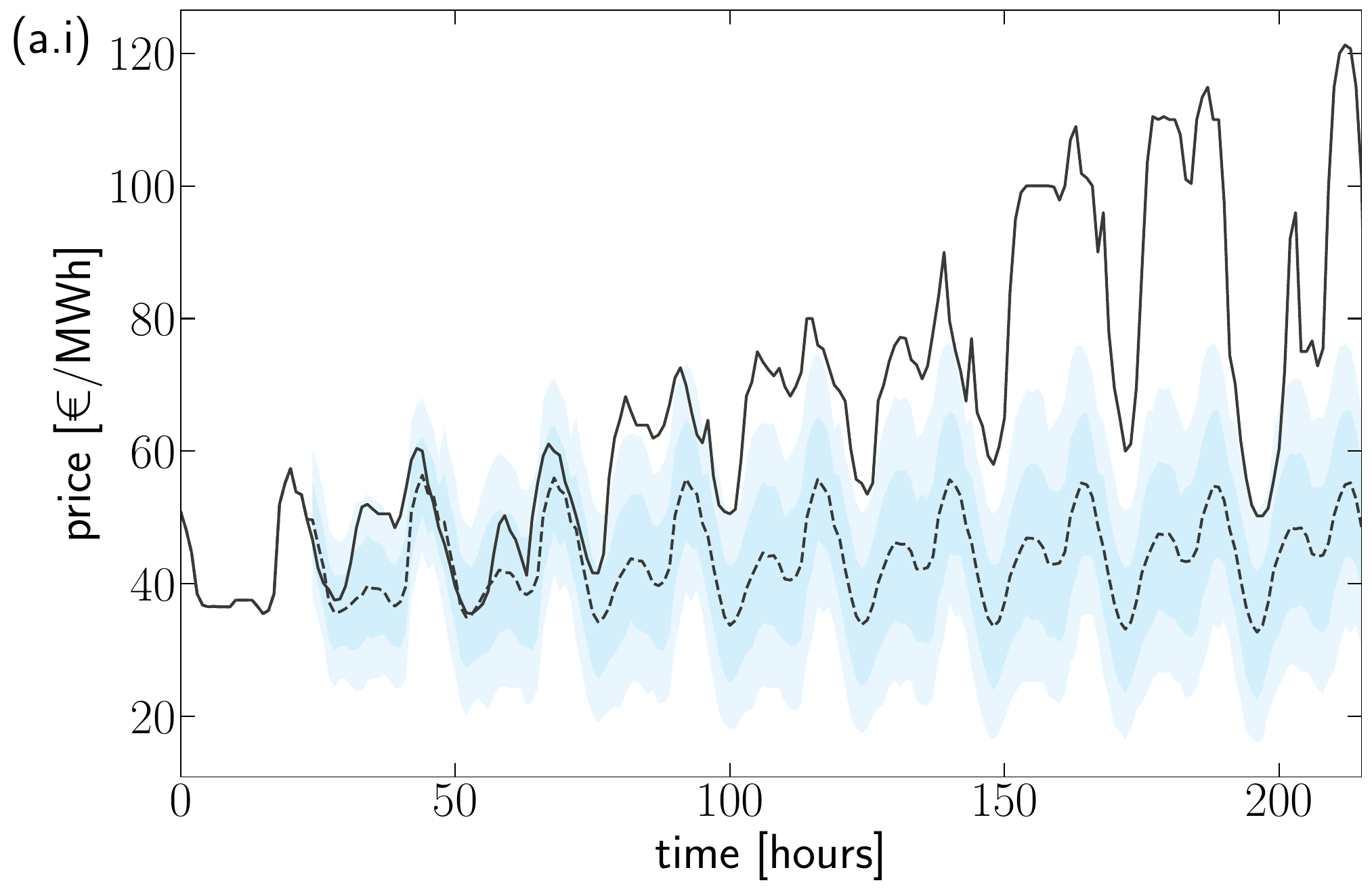}
\end{subfigure}
\hfill
\begin{subfigure}[t]{0.35\textwidth}
\includegraphics[width=1\textwidth]{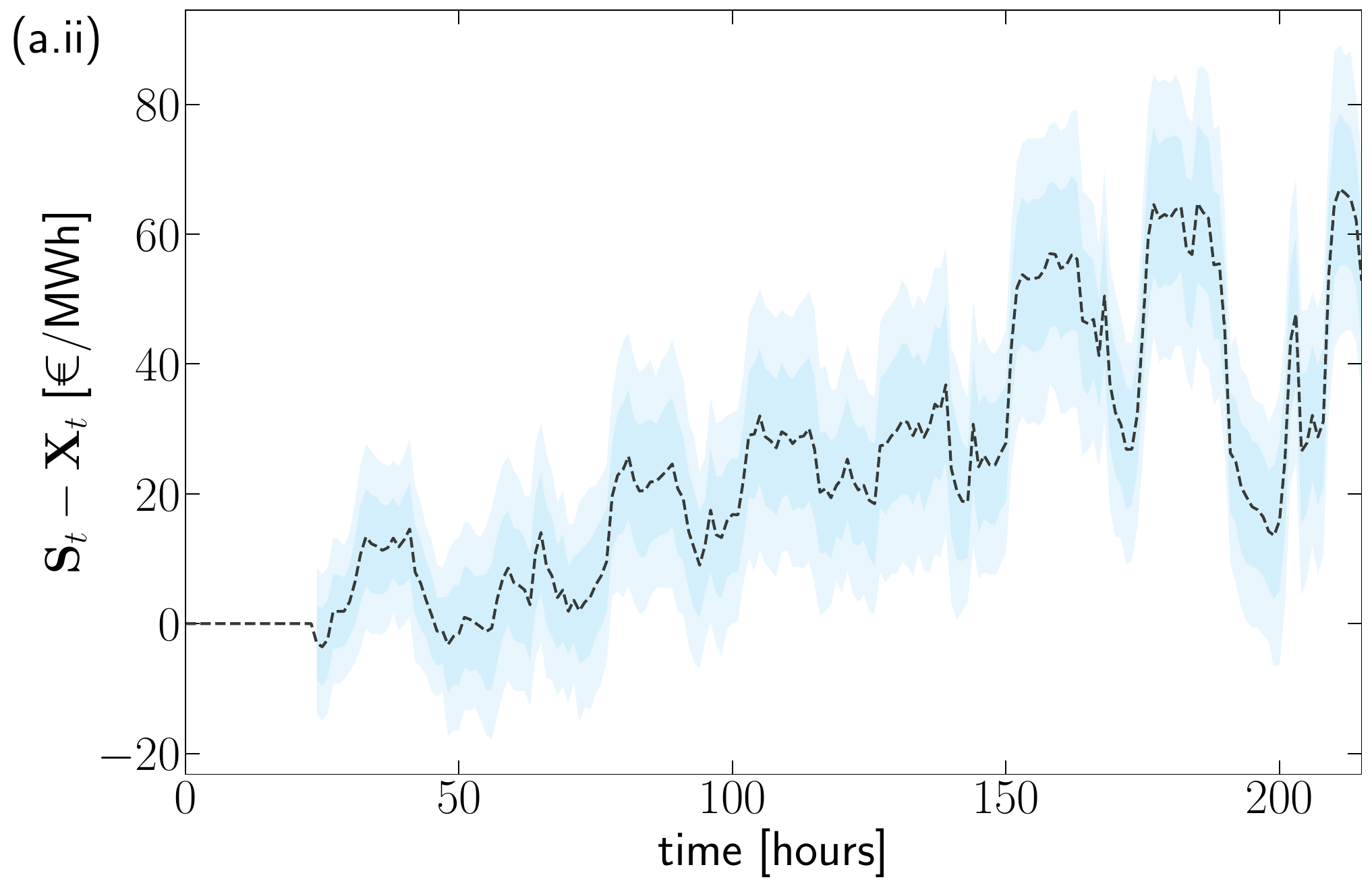}
\end{subfigure}
\hfill\mbox{}

\mbox{}\hfill
\begin{subfigure}[t]{0.35\textwidth}
\includegraphics[width=1\textwidth]{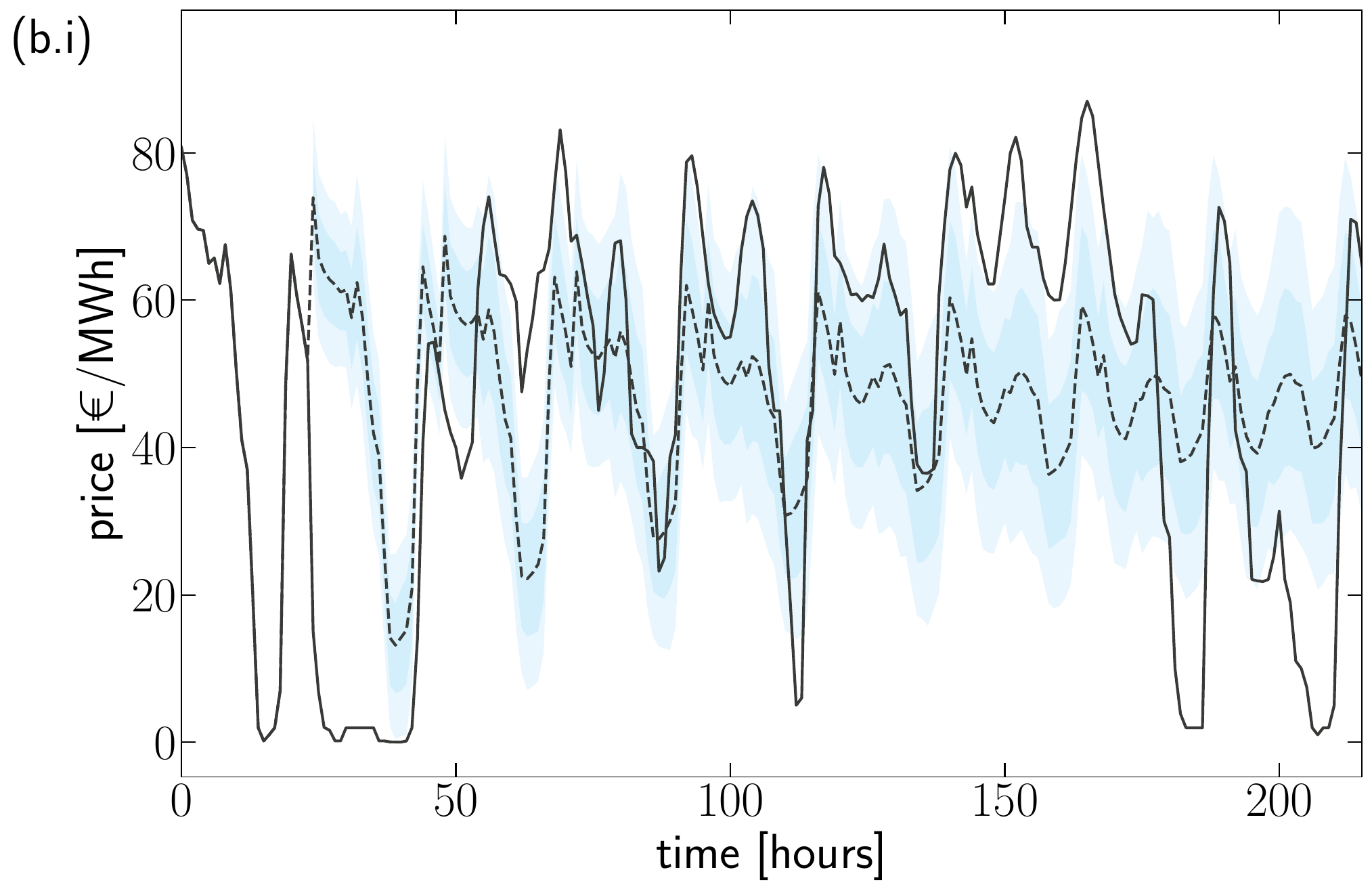}
\end{subfigure}
\hfill
\begin{subfigure}[t]{0.35\textwidth}
\includegraphics[width=1\textwidth]{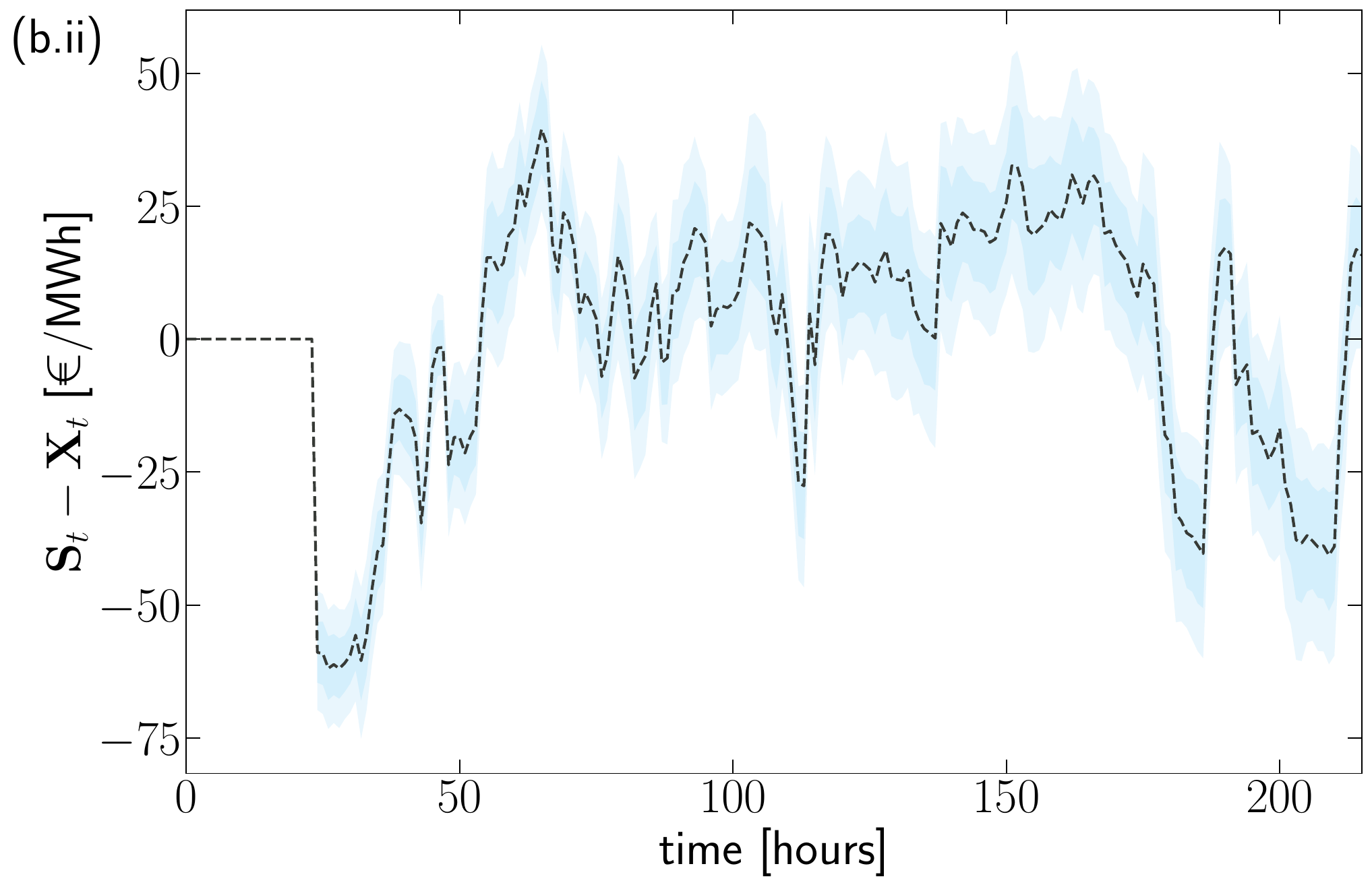}
\end{subfigure}
\hfill\mbox{}

\mbox{}\hfill
\begin{subfigure}[t]{0.35\textwidth}
\includegraphics[width=1\textwidth]{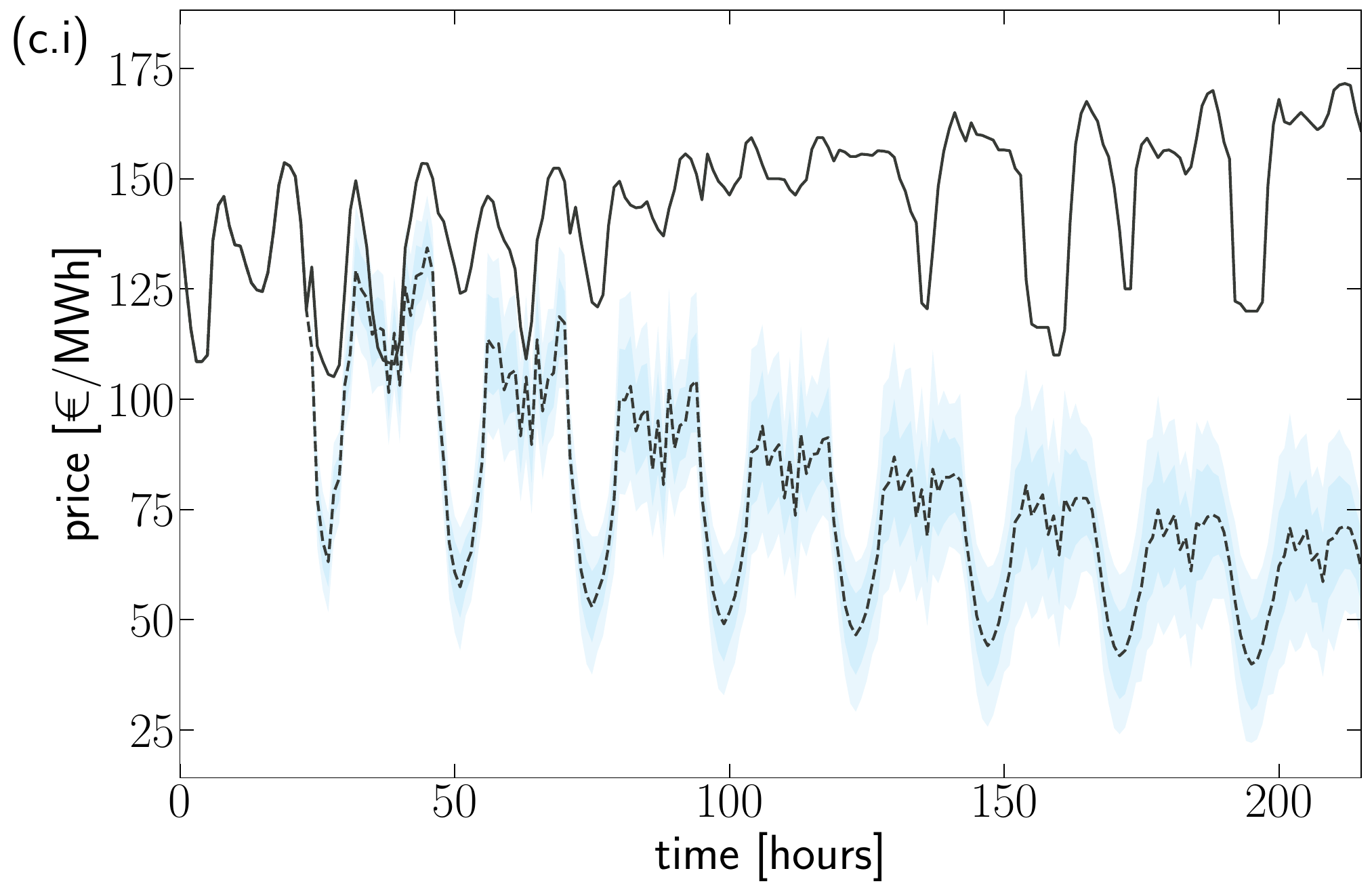}
\end{subfigure}
\hfill
\begin{subfigure}[t]{0.35\textwidth}
\includegraphics[width=1\textwidth]{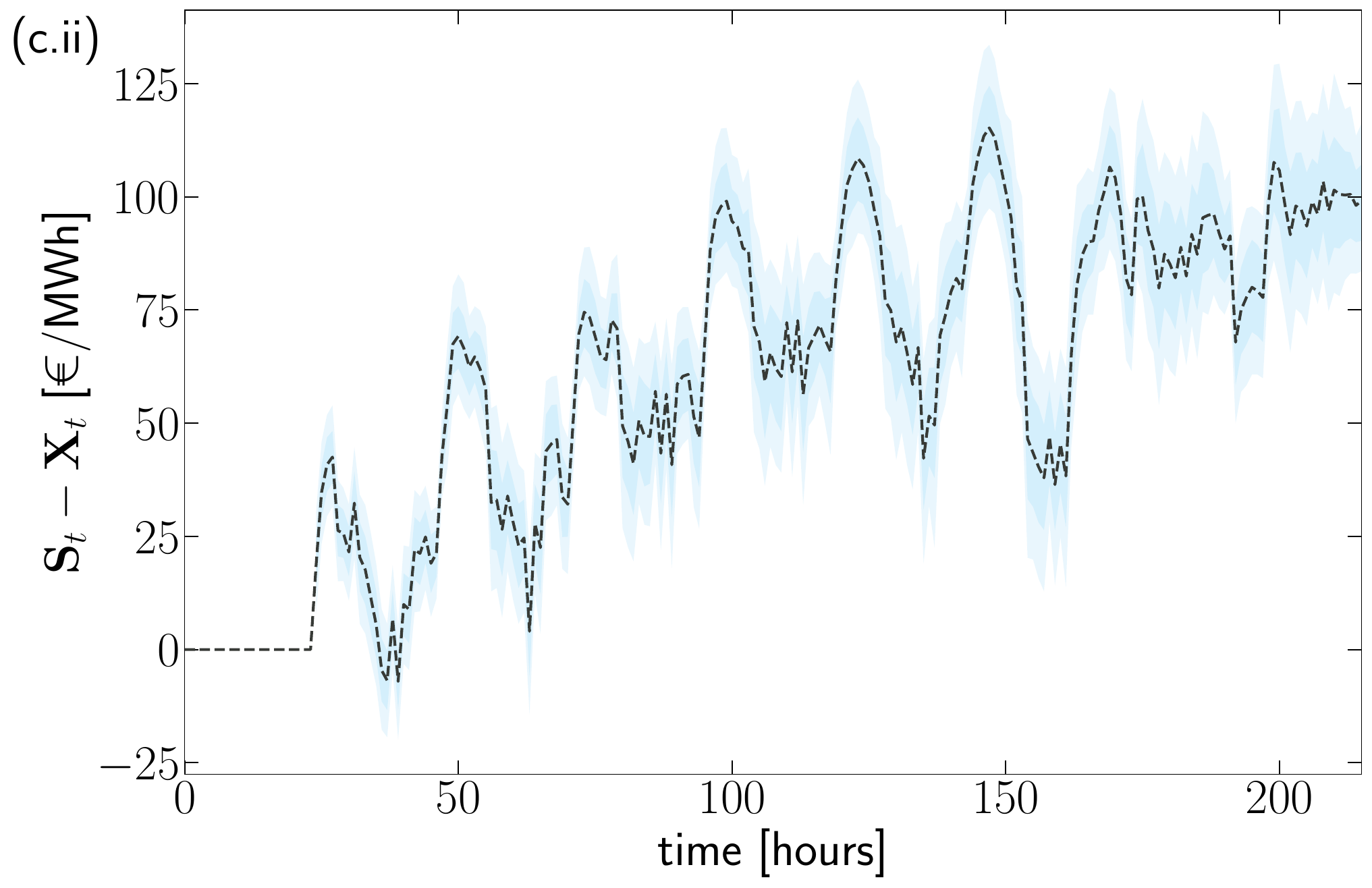}
\end{subfigure}
\hfill\mbox{}

\caption{Assessment of $\mathbf{X}_t$ obtained from the LE in non-stationary
conditions.
Each row corresponds to a different scenario with initial dates:
(a) 1/1/2021, (b) 8/5/2021, and (c) 6/9/2021.
Left column: comparison between the true electricity price, $\mathbf{S}_t$,
(solid line) and $10^3$ simulated price paths of $\mathbf{X}_t$.
Right column: time evolution of $\mathbf{S}_t - \mathbf{X}_t$.
Dashed lines correspond to the mean of $\mathbf{X}_t$ (left column) and mean of
$\mathbf{S}_t - \mathbf{X}_t$ (right column) over all simulated paths.
Shaded areas in both columns delimit the same percentile ranges as in Fig.~\ref{fig:stationary-gle}.}
\label{fig:gle-validation-scenarios}
\end{figure*}

\begin{figure*}[t]
\centering
\captionsetup{justification=raggedright}
{\includegraphics[width=\textwidth]{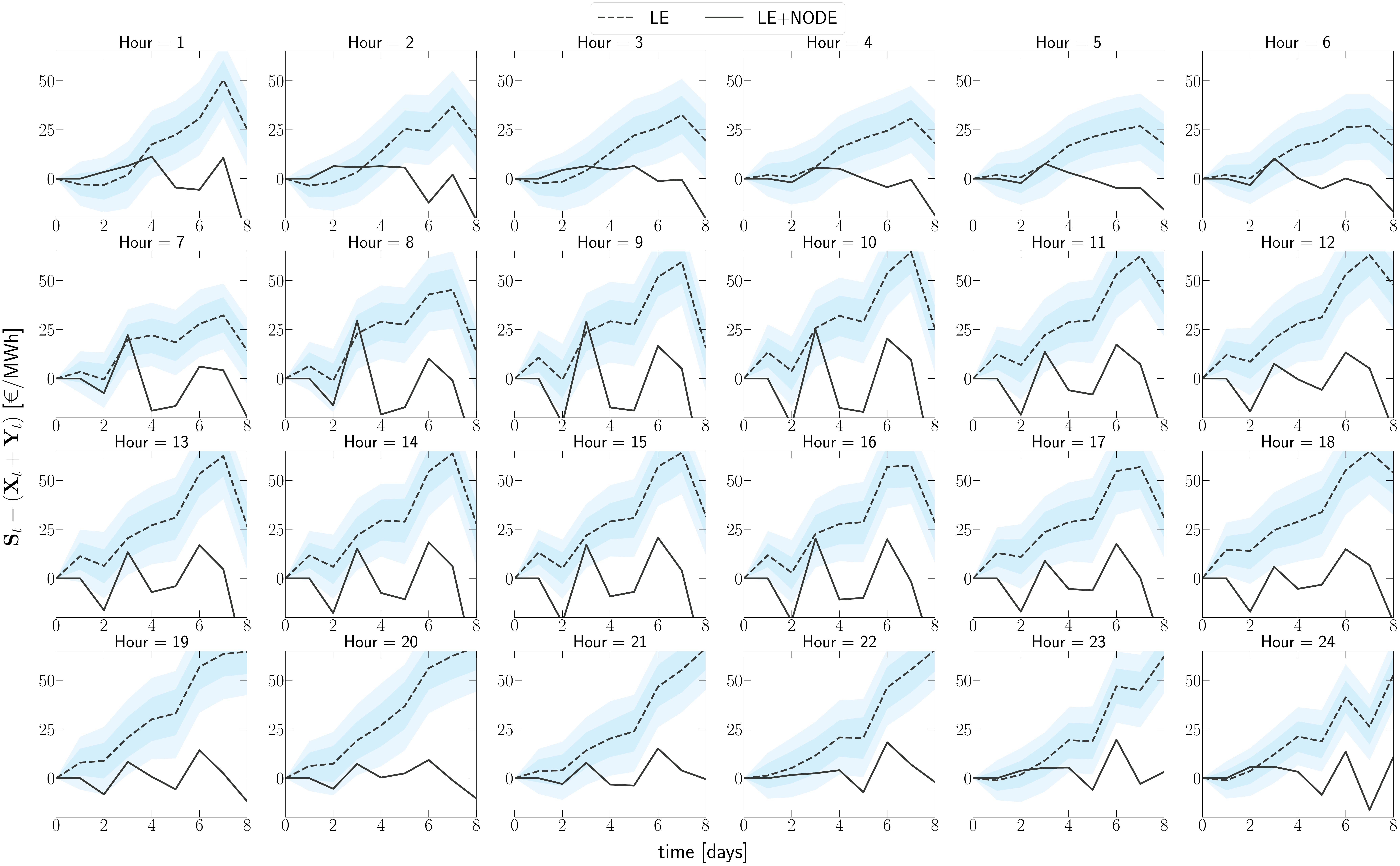}}
\caption{Hourly training dataset $\mathcal{D}^{p}_{n}, \text{with } p=9, \, n=10^3$, of scenario (a) in
Fig.~\ref{fig:gle-validation-scenarios}
computed as the difference between the true electricity price, $\mathbf{S}_t$,
and the simulated paths of  $\mathbf{X}_t$ obtained from the LE.
Dashed lines correspond to the mean of $\mathbf{S}_t - \mathbf{X}_t$.
Shaded areas delimit the same percentile ranges as in Fig.~\ref{fig:stationary-gle}.
These trajectories are equivalent to the time evolution depicted in Fig.~\ref{fig:gle-validation-scenarios} (a.ii) but rearranged into the 24 hourly dimensions considered in the LE.
Solid lines correspond to the mean of $\mathbf{S}_t - (\mathbf{X}_t + \mathbf{Y}_t)$, i.e., the error between the true price and the out-of-sample prediction of the
LE and the NODE.}
\label{fig:training-set}
\end{figure*}

The results of Fig.~\ref{fig:stationary-gle} corroborate the LE's effectiveness
in generating a robust approximation of the time evolution of the Spanish electricity
price under stationary conditions when time-dependent
effects are absent.
This validation of the price representation of the LE in stationary conditions
provides valuable insights into the short-term dynamics driving the Spanish
day-ahead market.
Among these dynamics, we can identify the daily price fluctuations,
the mean-reversion effects across different hours, and the distinct equilibrium
prices that exist for each hour of the day.
These short-term characteristics arise from the estimated
drift and diffusion terms from Eq.~\eqref{eq:km-definitions} that govern
Eq.~\eqref{eq:gle}.

To further assess the performance of the LE, we report in
Fig.~\ref{fig:gle-validation-scenarios} a comparison between the true
electricity-price signal and the simulated price paths generated by the LE
in various non-stationary scenarios.
Subplots (a.i) and (c.i) illustrate an upward trend in the electricity price
that the LE is unable to replicate.
In scenario (a), the initial condition is close to the equilibrium
prices.
Hence, the simulated paths remain within the same initial price ranges.
On the other hand, in scenario (c), the initial condition is far from the
equilibrium prices and therefore the simulated paths tend to revert to the
equilibrium values, amplifying the price forecasting error of the LE as is
evident from subplot (c.ii).
In contrast, subplot (b.i) represents a scenario characterized by a
prominent volatility.
The initial condition for scenario (b) exhibits a substantial price
variation, with many hourly prices deviating from the equilibrium ones.
As the LE reverts the initial prices toward the equilibrium values,
the large volatility is dampened, yielding a poor predictive
performance in this high volatility scenario.
In summary, these scenarios underscore the inability of the LE formulation
to anticipate non-stationary features, and the necessity of implementing the NODE.

\begin{figure*}
\centering
\captionsetup{justification=raggedright}

\mbox{}\hfill
\begin{subfigure}[t]{0.35\textwidth}
\includegraphics[width=\textwidth]{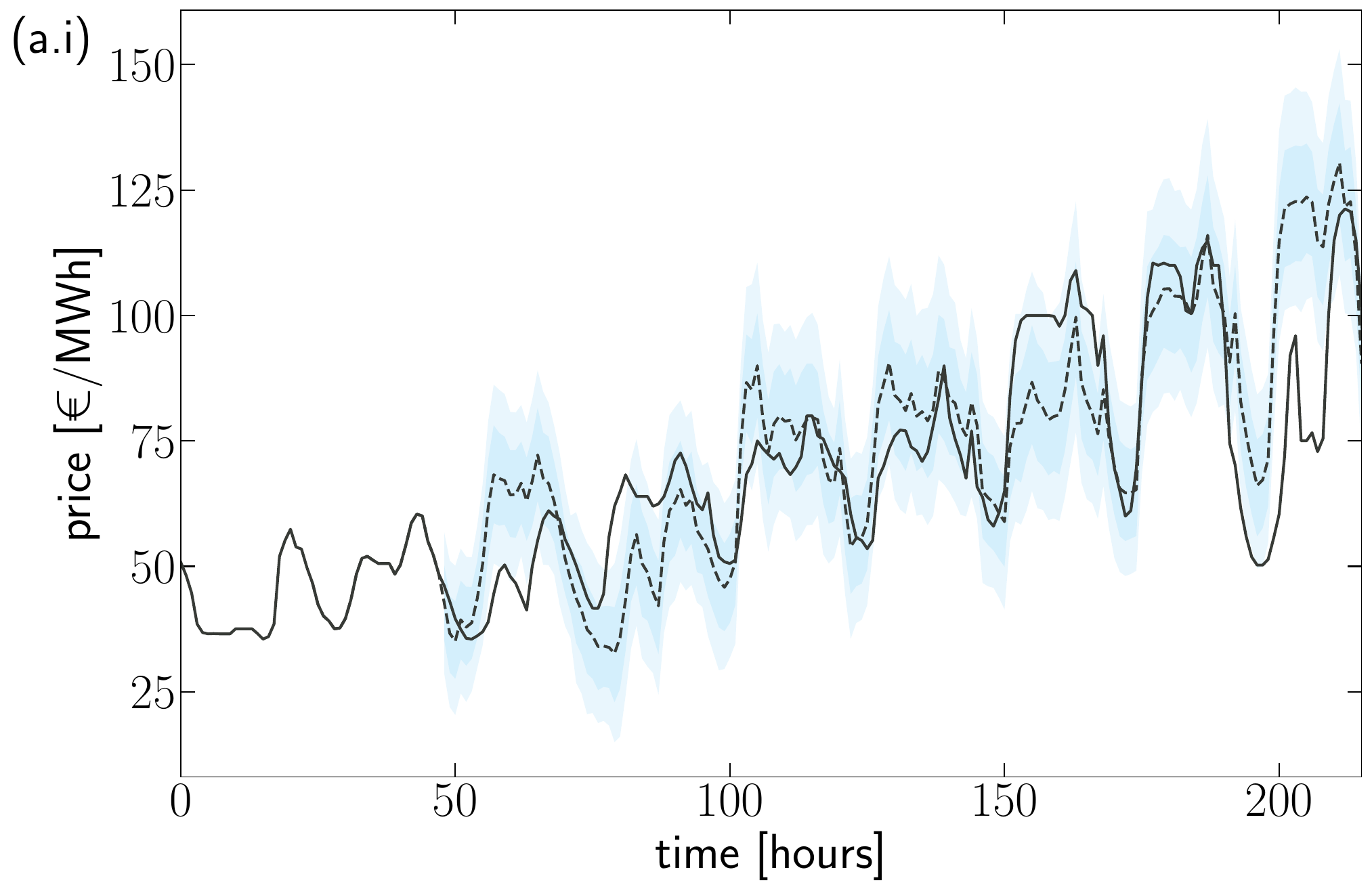}
\end{subfigure}
\hfill
\begin{subfigure}[t]{0.35\textwidth}
\includegraphics[width=\textwidth]{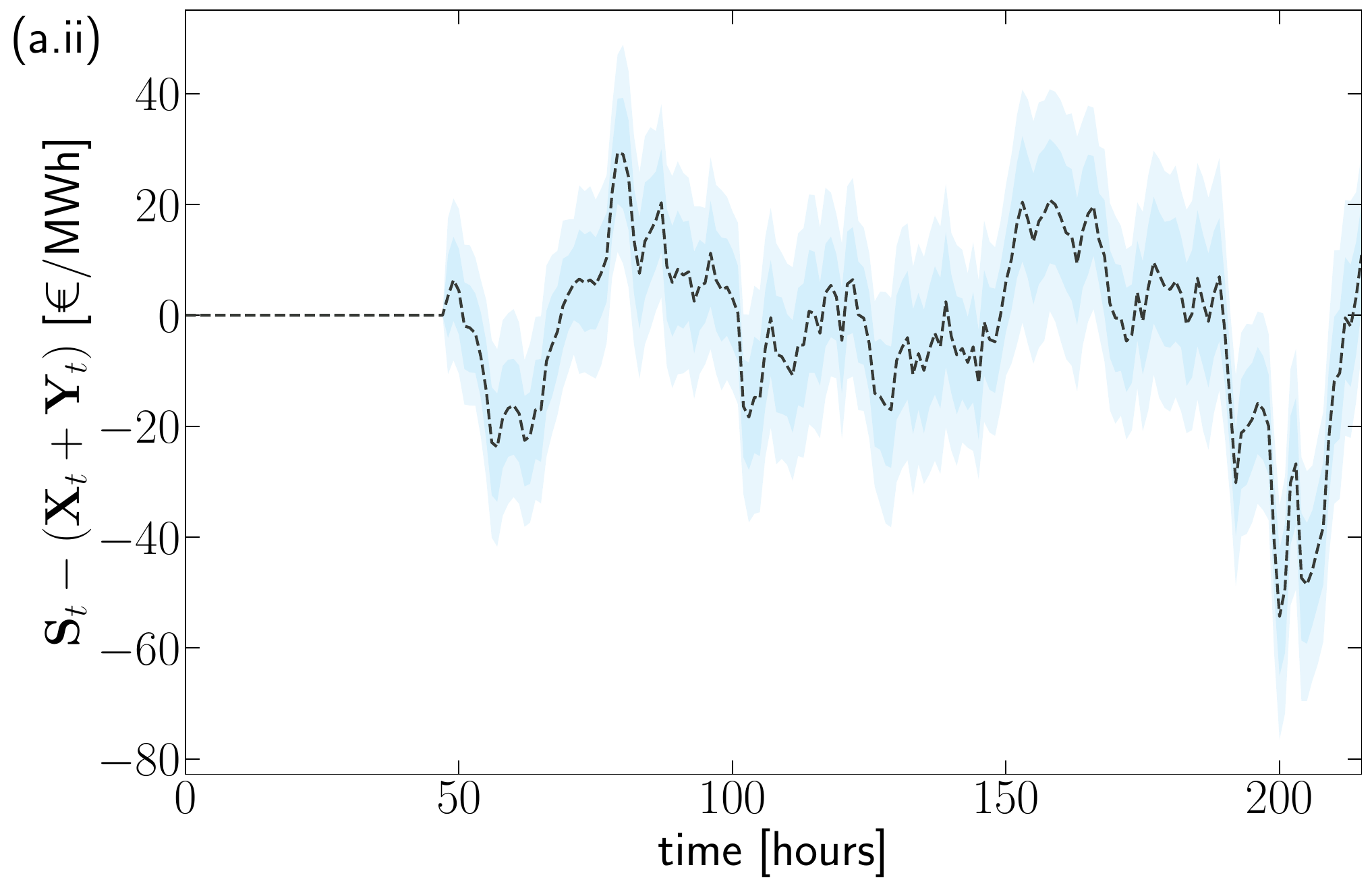}
\end{subfigure}
\hfill\mbox{}

\mbox{}\hfill
\begin{subfigure}[t]{0.35\textwidth}
\includegraphics[width=1\textwidth]{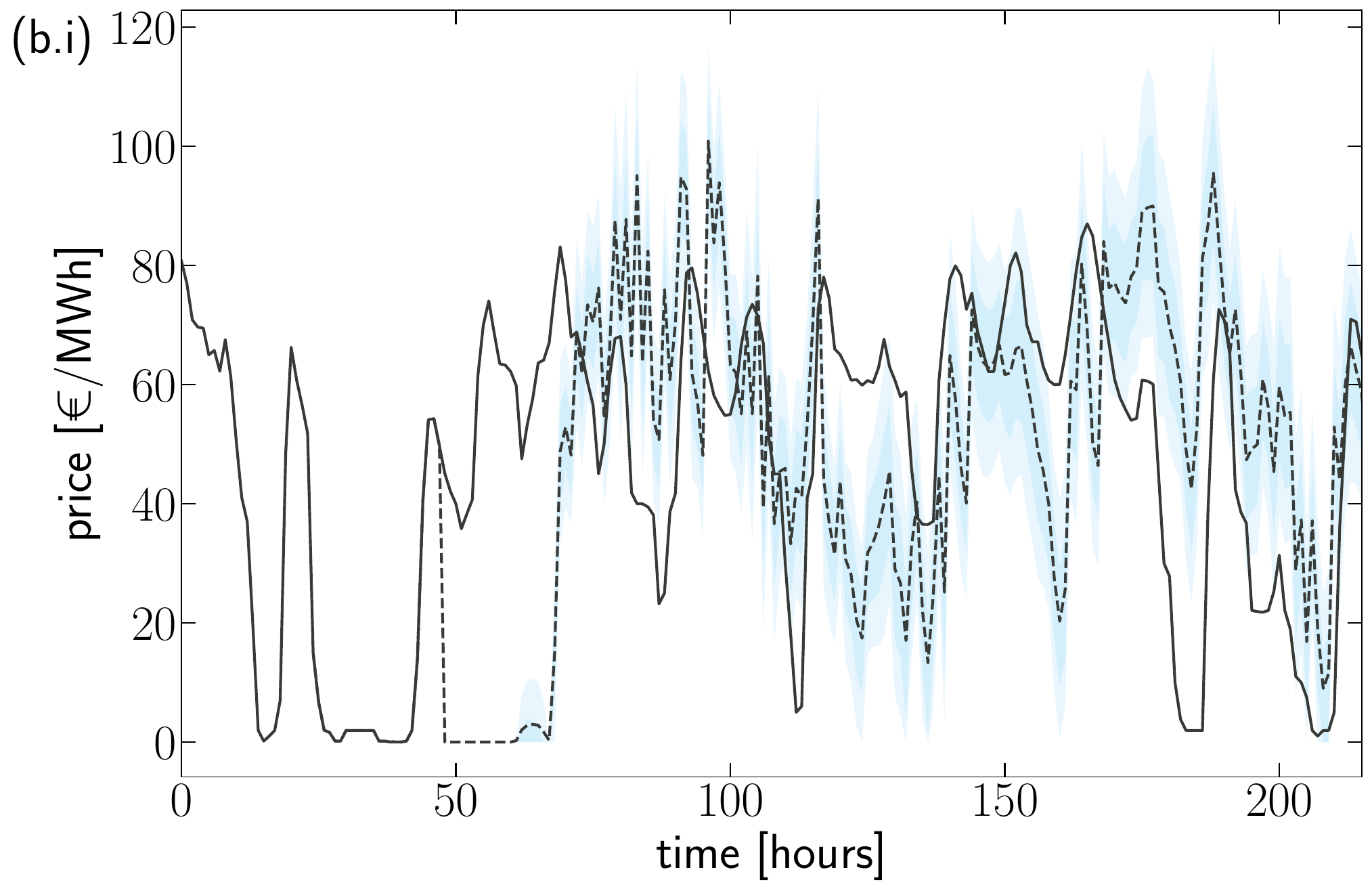}
\end{subfigure}
\hfill
\begin{subfigure}[t]{0.35\textwidth}
\includegraphics[width=1\textwidth]{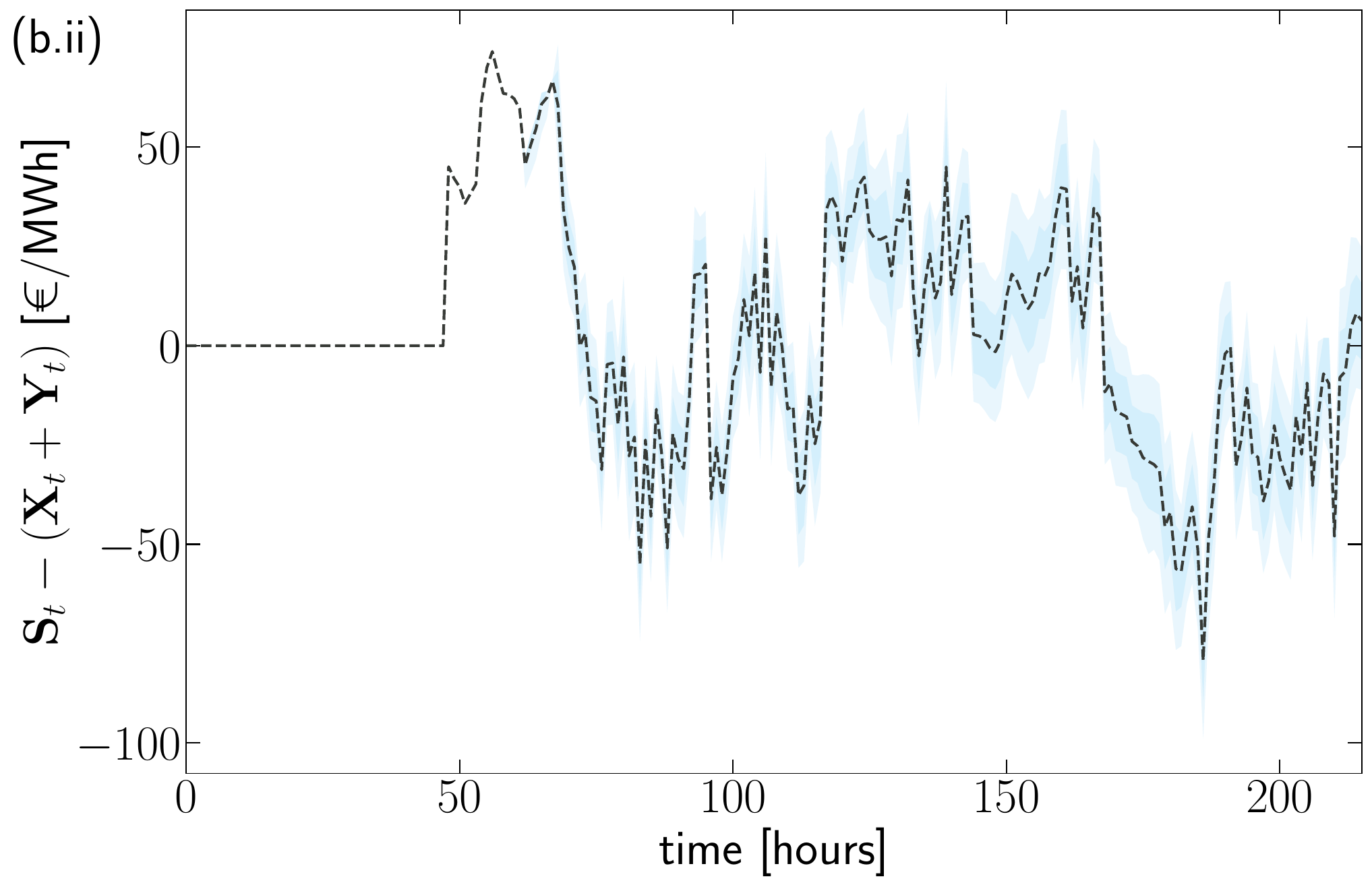}
\end{subfigure}
\hfill\mbox{}

\mbox{}\hfill
\begin{subfigure}[t]{0.35\textwidth}
\includegraphics[width=1\textwidth]{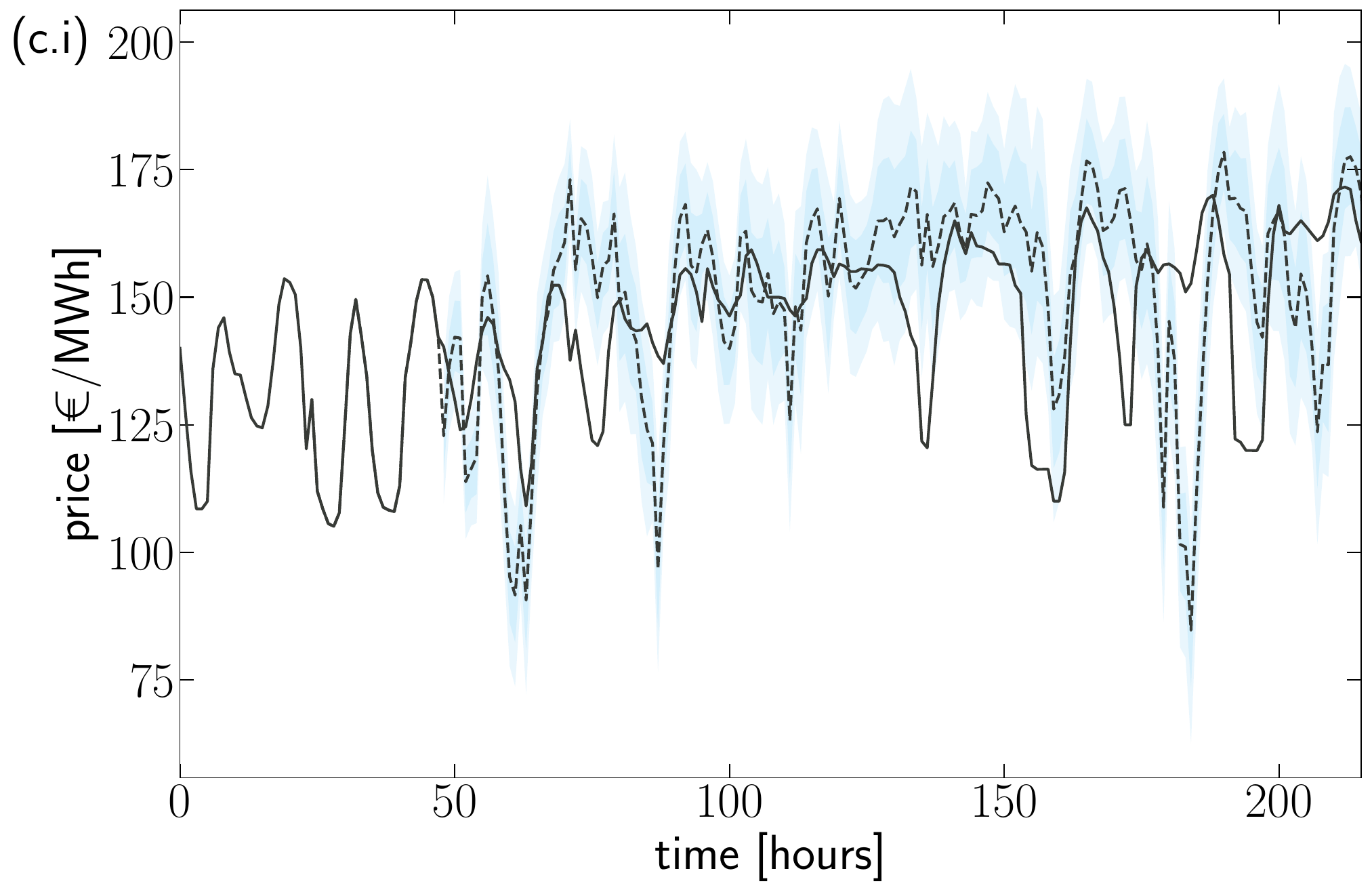}
\end{subfigure}
\hfill
\begin{subfigure}[t]{0.35\textwidth}
\includegraphics[width=1\textwidth]{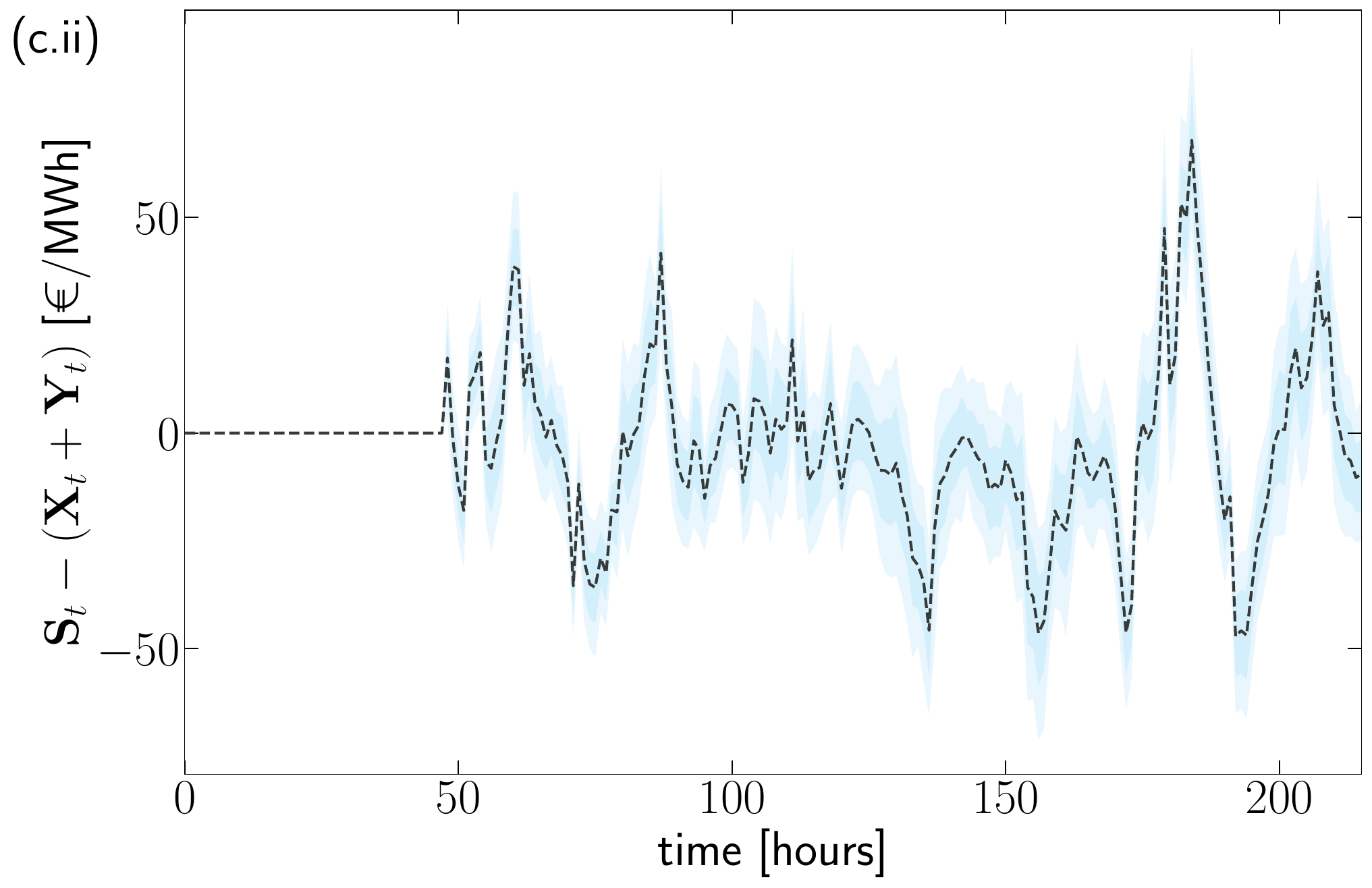}
\end{subfigure}
\hfill\mbox{}

\caption{Assessment of $\mathbf{X}_t + \mathbf{Y}_t$ obtained from the combined LE-NODE.
Each row corresponds to a different scenario with initial dates:
(a) 1/1/2021, (b) 8/5/2021, and (c) 6/9/2021.
Left column: comparison between the true electricity price, $\mathbf{S}_t$
(solid line) and the prediction of $\mathbf{X}_t + \mathbf{Y}_t$.
Right column: time evolution of $\mathbf{S}_t - (\mathbf{X}_{t} + \mathbf{Y}_t)$.
The first 48 h ($p = \{0, 1\}$) correspond to training samples for the NODE.
The remaining hours ($p = \{2, \ldots, 8\}$) contain out-of-sample predictions ($q=1$) of $\mathbf{Y}_t$ from the
NODE combined with the $n=10^3$ simulated paths of $\mathbf{X}_t$ generated by the LE.
Dashed lines correspond to the mean of $\mathbf{X}_t + \mathbf{Y}_t$ (left column) and mean $\mathbf{S}_t - (\mathbf{X}_t + \mathbf{Y}_t)$ (right column) over all simulated paths.
Shaded areas in both columns delimit the same percentile ranges as in Fig.~\ref{fig:stationary-gle}.}
\label{fig:prediction-scenarios}
\end{figure*}

\begin{figure*}
\centering
\captionsetup{justification=raggedright}
{\includegraphics[width=\textwidth]{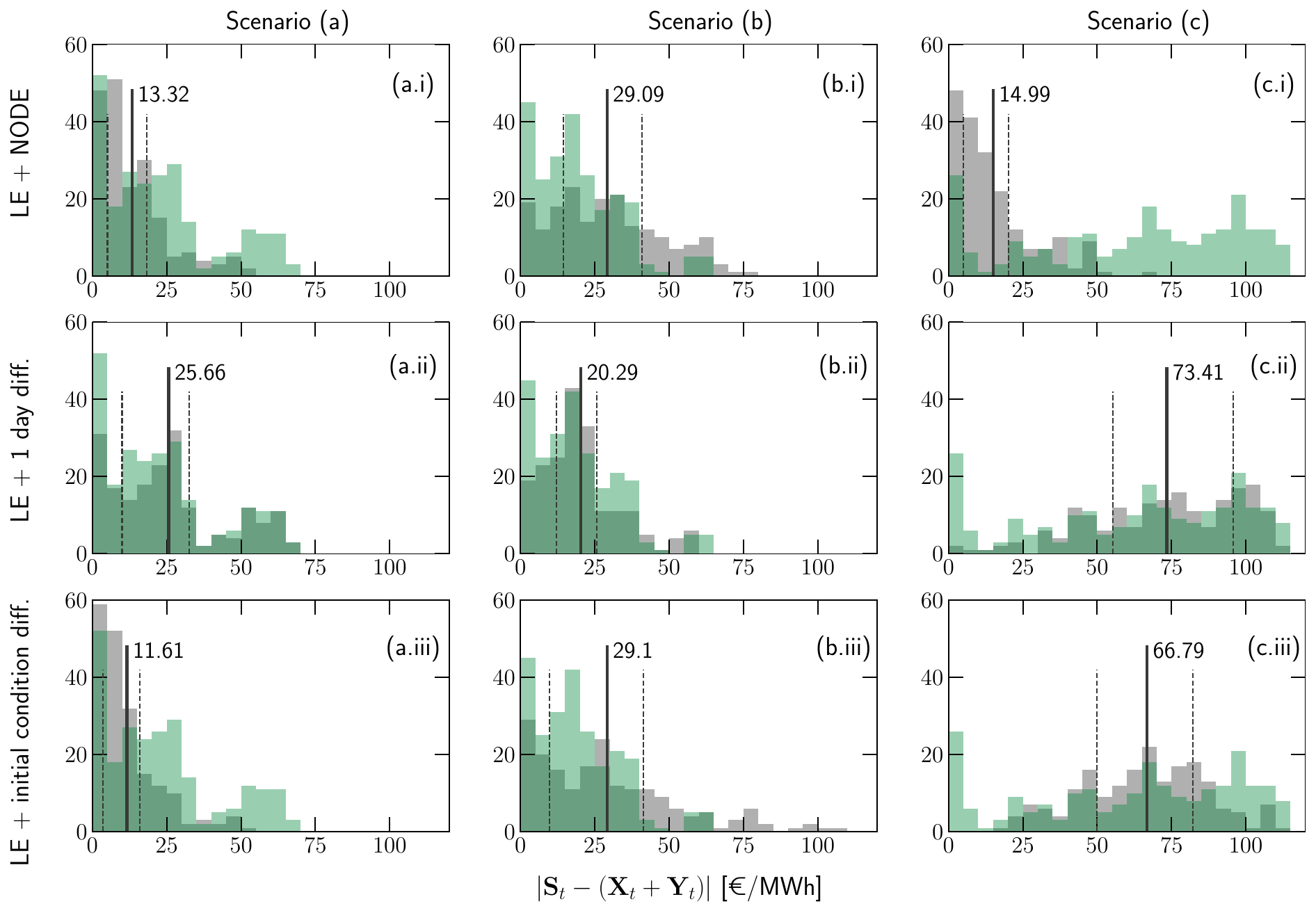}}
\caption{Performance histograms for the different approaches (specified on the
vertical axis), under the different scenarios (a), (b), and (c) of Fig.~\ref{fig:gle-validation-scenarios}.
Cases considered: LE + NODE (first row), LE + 1 day difference (second row), and
LE + initial condition difference (third row).
Performance metric: absolute error between price observations and predictions both for
stationary, only the LE model (green), and non-stationary models (gray) (note that the overlap of
the green and gray colors makes the former darker).
Solid lines indicate the MAE of the non-stationary prediction.
Dashed lines delimit the interquartile range of the non-stationary prediction.}
\label{fig:histogram-scenarios-methods}
\end{figure*}

\subsection{Extension with NODEs}
\label{subsec:neural-ode-technique}

The proposed NN architecture within the NODE consists of a
feed forward fully-connected network with one hidden layer.
The input and output dimension is 24, one per hourly difference, to maintain
the multivariate structure of the LE.
The hidden layer contains 96 neurons with hyperbolic tangent as the
activation function.
The time evolution, denoted as $t$ in Eq.~\eqref{eq:nural-ode}, is
measured in days.

The training process proceeds as follows:
at the initialization stage, we compute $\mathcal{D}^{p}_{n}$ using the
initial condition $\mathbf{Y}_{t_0} = \mathbf{S}_{t_0} - \mathbf{X}_{t_0}$,
where $t_0$ is the initial time step.
We note that $\mathbf{Y}_{t_0} = \mathbf{0}$, being the initial condition
$\mathbf{X}_{t_0} = \mathbf{S}_{t_0}$.
The training objective for the NODE is then to minimize the disparity between
the real difference trajectory, $ \{ \mathbf{S}_{t_0+1} - \mathbf{X}_{t_0+1},
\, \ldots, \, \mathbf{S}_{t_0+p} - \mathbf{X}_{t_0+p} \}$, and the predicted
one.
We choose the mean absolute error (MAE) as the loss function to be minimized,
and the root mean squared propagation as the optimizer algorithm to update
the NN weights, $\theta$.
During the training process, we employ a mini-batch gradient descent
algorithm with 32 samples per batch and $2 \times 10^3$ epochs with a
learning rate of $10^{-3}$.
We implement and train the NODE using the Python library
\textit{torchdiffeq}.~\cite{torchdiffeq}
It is worth emphasizing that we have selected empirically all
hyper-parameters related to the NN architecture and optimization scheme based
on the NODE performance.

Figure~\ref{fig:training-set} illustrates the training dataset
$\mathcal{D}^{p}_{n}$ for scenario (a) of Fig.~\ref{fig:gle-validation-scenarios}.
Each hourly subplot depicts the trajectory of the differences between the
true price, $\mathbf{S}_t$, and the mean over all simulated paths generated
by the LE (dashed line).
The NODE attempts to fit the empirical differences, reconstructing the dashed-line trajectory.
Our training process, involving the epochs and mini-batch procedure described
earlier, continually feeds the NODE with a random subset of price differences'
trajectories.
This randomized strategy ensures that, on average, the NODE approximates very
well the dashed-line trajectory, reducing the error between $\mathbf{S}_t$ and our framework, $\mathbf{X}_t + \mathbf{Y}_t$, close to 0, as evidenced by the solid lines.

To assess the predictive performance of our framework, in general, and the NODE,
in particular, we undertake the following validation procedure for each scenario
represented in Fig.~\ref{fig:gle-validation-scenarios}:
using $\mathcal{D}^{p}_{n}$, $p = \{1, \ldots, 8\}$, we train the NODE up to time step $p$ and predict the next
set of time steps, $p+q$, corresponding to the out-of-sample predictions of
$\mathbf{Y}_{t_0+p+q}$.
For the sake of efficiency, we adopt the smallest possible validation time
step, i.e., $q=1$ day.
We combine $\mathbf{Y}_{t_0+p+1}$ with the simulated price paths of $\mathbf{X}_{t_0+p+1}$ to generate $\mathcal{D}_{n}^{p, 1}$, and compare it
to $\mathbf{S}_{t_0+p+1}$.

Figure~\ref{fig:prediction-scenarios} plots $\mathbf{S}_t$ and the time
evolution of the electricity price for $\mathcal{D}^{p,1}_{n}, \,$ $p=\{1,
\ldots, 8\}$.
Comparing these results with the baseline price approximations of the LE,
shown in Fig.~\ref{fig:gle-validation-scenarios}, we observe a notable
improvement in the price reconstruction due to the predictions of
$\mathbf{Y}_t$ generated by the NODE.
These predictions substantially enhance the representation of the
electricity-price dynamics, reducing the mismatch between the actual
electricity prices and the simulated price paths generated by the LE.
However, the accuracy of these predictions can vary across different
scenarios.
When there is a persistent trend in the electricity day-ahead market, such as
in scenarios (a) and (c), the NODE accounts for this external effect
efficiently.
This leads to compelling forecasts of the time-dependent
component, $\mathbf{Y}_t$, of the electricity day-ahead prices.
Remarkably, scenario (c) presents a dynamic interplay between the LE and the
NODE.
With an initial condition located far from equilibrium prices,
the LE drives the electricity prices toward the equilibrium values,
as seen in Fig.~\ref{fig:gle-validation-scenarios} (c.i).
Nevertheless, the existing upward trend cancels completely the historical
mean-reversion effect, maintaining the actual electricity prices even further
from their historical expected values.
The NODE determines this change in the dynamic regime and exerts a correcting
force to counteract the mean-reversion feature of the LE.
Finally, the increased volatility in scenario (b) presents a challenging
condition for the NODE, resulting in a low prediction accuracy.
The rapid price variations in scenario (b) likely contribute to the reduced
performance of the NODE, as opposed to scenarios (a) and (c), where the
upward trend presents a smooth and sustained external effect that the NODE
can capture effectively.

\subsection{Benchmarking}
\label{subsec:benchmarking}

Figure~\ref{fig:histogram-scenarios-methods} presents the histograms of the
absolute differences between the actual prices, $\mathbf{S}_t$, and the
expected values of the out-of-sample predictions, $\mathbf{X}_t +
\mathbf{Y}_t$.
The histograms facilitate a comparative analysis of the outcomes generated by
our framework and the na\"ive methods detailed in
Sec.~\ref{subsec:naive-methods} across each scenario of Figs.~\ref{fig:gle-validation-scenarios} and \ref{fig:prediction-scenarios}.
Furthermore, we calculate two key metrics for each method and scenario:
the MAE (vertical solid line) and interquartile range (vertical dashed lines).
Our proposed framework, the synergistic combination of LE and NODE,
clearly outperforms the na\"ive methods in scenario (c), and consistently
emerges as the most robust technique across all scenarios due to its
remarkable trade-off between the MAE and the interquartile range.
In scenario (a), the combined LE-NODE (a.i) reveals a similar performance to
the ``LE + initial condition difference"  (a.iii) approach, which proves to
be the best model.
However, the ``LE + 1 day difference" method yields the lowest MAE score in
scenario (b), as is evident from subplot (b.ii).
In this scenario (b), our framework (b.i) and the ``LE + initial condition difference"
(b.iii) present identical MAEs, but our framework exhibits a lower interquartile
range.
It is worth noting that our framework exhibits its lowest accuracy in this
high volatility scenario (b), as previously discussed in relation to
Fig.~\ref{fig:prediction-scenarios}~(b.i) and (b.ii).
Conversely, the na\"ive methods display higher error rates and a larger
dispersion in scenario (c), observe (c.ii) and (c.iii), compared to the
combined LE-NODE methodology (c.i).
This further highlights the superiority of the combined LE-NODE.

\section{Conclusions}
\label{sec:conclusions}

We have introduced a rational and systematic data-driven
mathematical framework to analyze non-stationary time series.
The proposed framework is applied on electricity-price series, a topical
subject due to the recent energy crisis worldwide, and addresses
simultaneously stationary and non-stationary features of such series,
advancing the non-stationary electricity-price modeling arena.
Because of its generality and versatility, our framework can accommodate time
series exhibiting concurrent stationary and non-stationary features in other
areas and application domains.

It integrates synergistically the LE with a NODE to forecast the
short-term time evolution of the electricity price following a two-stage
approach.
First, the drift and diffusion terms of the LE are fitted to historical
electricity prices in order to yield a reliable price formulation.
However, the expected prices lie at the heart of the estimation of the
drift-diffusion terms, and the LE necessarily accounts for the stationary
behavior of the electricity price only.
We subsequently employ a NODE to learn the difference between the actual
electricity-price time series and the simulated prices generated by the LE.
By learning this difference, the NODE captures the underlying dynamics of the
non-stationary components of the price behavior that the LE cannot
adequately approximate.
Therefore, the NODE nicely complements the LE and extends the formulation to
address effectively both stationary and non-stationary electricity-price time
series.
We can then infer the (short-term) dynamic laws dictating the electricity
price evolution in stationary conditions, while at the same time account for
external effects.

Our study uses the Spanish electricity day-ahead prices as a prototypical
system to showcase the applicability of our methodology to real complex
systems.
The results reveal that the LE formulation successfully unravels
electricity-price features such as the mean-reversion effect, the existing equilibrium prices for each hour,
and the daily fluctuations.
However, the LE itself is not sufficient to accurately predict the
electricity prices, particularly when persistent trends or changes in
volatility occur in the day-ahead market.
To overcome this limitation, we integrate the LE with a NODE. Specifically,
the output of the NODE corrects the deviation between the actual price-time
series and the price representation obtained from the LE, resulting in a
substantially more precise electricity-price forecast.
Furthermore, to assess the performance of our methodology, we conduct a
comparative analysis with a number of na\"ive methods, showing that our model
is reliable and robust.
In detail, the comparison illustrates similar price approximations between
our model and the na\"ive methods across two non-stationary scenarios: one with
a positive trend starting from equilibrium prices and another with increased
volatility levels.
Nevertheless, the LE-NODE framework outperforms the na\"ive methods when there
is an upward external drift originating from price values far from historical
equilibrium prices.
This significant outcome is because the NODE counteracts efficiently the
price dynamics dictated by the LE, i.e., the mean-reversion effect, which is
temporarily not applicable in this external drift scenario.

This fusion of the LE with a NODE unravels the price dynamics in the
stationary regime and provides satisfactory forecasts under non-stationary
effects.
Specifically, it identifies and predicts first-moment variations in the
electricity-price signal, such as upward trends.
However, the present NODE, as we have designed it, will not yield accurate
results when dealing with second-moment variations, e.g., changes in price
volatility, as in scenario (b) of Fig.~\ref{fig:prediction-scenarios}.
Consequently, future refinements will aim at enhancing the proposed framework to
account carefully for varying volatility.
This might entail exploring deep, convolutional, or recurrent NN architectures
within the NODE and/or considering the adoption of neural stochastic differential
equations~\cite{Kidger2021b, Oleary2022} as an alternative to the NODE.
Another worthwhile line of enquiry would be to further enhance the learning
process by introducing elements of Bayesian inference, as in our recent
studies in Refs~\onlinecite{Yatsyshin2022,Malpica-Morales2023}, to enable
uncertainty quantification which is not native to NNs.
Finally, the NODE designed within our framework may be substituted by standard non-stationary techniques, such as empirical mode decomposition,~\cite{Lahmiri2017}  wavelets,~\cite{Conejo2005} or autoregressive integrated moving average models.
Yet, the implementation of these standard techniques should be carefully
evaluated and benchmarked with our NODE which combines superior data efficiency
with simplicity. 
We shall examine these and related questions in future
studies.

\section*{ACKNOWLEDGMENTS}

A.M. was supported by the Imperial College London President’s PhD Scholarship
scheme. S.K. was supported by the ERC-EPSRC Frontier Research Guarantee
through Grant No. EP/X038645, ERC through Advanced Grant No. 247031, and
EPSRC through Grants No. EP/L025159 and EP/L020564.

\appendix

\section{Estimation of the drift and diffusion coefficients}
\label{app:drift-diffusion-estimation}

The drift and diffusion coefficients of the LE in Eq.~\eqref{eq:gle} can be computed using the definitions for the Kramers-Moyal expansion coefficients,
\begin{eqnarray}
\mu^{h_i}(\mathbf{X}) &&= \lim\limits_{\tau \rightarrow 0} \frac{\langle X^{h_i}_{t+\tau} - X^{h_i} \rangle}{\tau} = D^{(1)}_{h_i}(\mathbf{X}) \nonumber
\\
\frac{1}{2} \sigma_{h_i h_k} (\mathbf{X}) \sigma_{h_j h_k} (\mathbf{X}) &&= \frac{1}{2} \lim\limits_{\tau \rightarrow 0} \frac{\langle [X^{h_i}_{t+\tau} - X^{h_i}] [X^{h_j}_{t+\tau} - X^{h_j}] \rangle}{\tau} \nonumber
\\
&&= D^{(2)}_{h_i h_j}(\mathbf{X}), \label{eq:km-definitions}
\end{eqnarray}
where $\langle \cdot \rangle$ is the shorthand notation for conditional
expectation, and $D^{(1)}, D^{(2)}$ represent the first and second
Kramers-Moyal coefficients, respectively.
We note that we adopted Einstein's notation for the diffusion coefficient.

The conditional expectation of Eq.~\eqref{eq:km-definitions}
requires the use of the historical PDFs that govern $D^{(1)}_{h_i}$
and $D^{(2)}_{h_ih_j}$.
These PDFs represent the joint probability of the random variables:
\begin{eqnarray}
P^{h_i} &&= (P^{h_i}_1, P^{h_i}_2) =  (X^{h_i}_{t}, X^{h_i}_{t+1} - X^{h_i}_{t}) \nonumber \\
P^{h_ih_j} &&= (P^{h_ih_j}_1, P^{h_ih_j}_2, P^{h_ih_j}_3 ) \nonumber \\
&&=  \left( X^{h_i}_{t}, X^{h_j}_{t}, \frac{1}{2} (X^{h_i}_{t+1} - X^{h_i}_{t}) (X^{h_j}_{t+1} - X^{h_j}_{t}) \right).
\label{eq:pdfs}
\end{eqnarray}
We approximate the PDFs of $P^{h_i}$ and $P^{h_ih_j}$
applying a kernel density estimation~\cite{Silverman1986} technique over the datasets:
\begin{eqnarray}
\mathcal{P}^{h_i} = &&\{ (x^{h_i}_{t+k}, \, x^{h_i}_{t+k+1} - x^{h_i}_{t+k}) \} \nonumber \\
\mathcal{P}^{h_ih_j} = &&\left\lbrace \left(x^{h_i}_{t+k}, \, x^{h_j}_{t+k}, \, \frac{1}{2} (x^{h_i}_{t+k+1} - x^{h_i}_{t+k}) (x^{h_i}_{t+k+1} - x^{h_i}_{t+k}) \right) \right\rbrace \nonumber \\
&& k = 0, \ldots, N-2,
\label{eq:pdf-datasets}
\end{eqnarray}
with $N$ being the number of available data samples.
We replace $\tau$ from Eq.~\eqref{eq:km-definitions} by $1$ in
Eqs.~\eqref{eq:pdfs} and \eqref{eq:pdf-datasets}, as $X^{h_i}_t$ evolves on a
daily basis in Eq.~\eqref{eq:gle}, with $\tau = 1$ being the minimum time
resolution that we can adopt.
We fit a single Gaussian kernel over each data sample of the datasets
$\mathcal{P}^{h_i}, \, \mathcal{P}^{h_ih_j}$.
The bandwidth $H$ of the Gaussian kernel follows Scott's
rule:~\cite{Scott1992}
\begin{equation}
H = (N^{-\frac{1}{d+4}})\mathbf{I}_{d},
\end{equation}
where $d$ indicates the input kernel dimensions ($d=2$ for $P^{h_i}$ and $d=3$
for $P^{h_ih_j}$) and $\mathbf{I}_d$ is a $d$-dimensional identity matrix.
The sum and normalization of all kernels for each
$\mathcal{P}^{h_i}, \, \mathcal{P}^{h_ih_j}$
yields an approximate reconstruction of the PDFs for $P^{h_i}$ and $P^{h_ih_j}$.
We then sample the approximated PDFs through a mesh resolution of ($10^3 \times 10^3$)
and ($10^2 \times 10^2 \times 5 \cdot 10^2$) datapoints for $P^{h_i}$ and $P^{h_ih_j}$, respectively.
Finally, we condition the obtained samples of the approximated PDF of $P^{h_i}$ on $X^{h_i}$ in order to calculate $D^{(1)}_{h_i}$,
\begin{equation}
D^{(1)}_{h_i}(\mathbf{X}) = \mathbb{E}[P^{h_i}_{2} | P^{h_i}_1 = X^{h_i}],
\end{equation}
where $\mathbb{E}$ represents the expected value.
Conversely, we assume that $D^{(2)}_{h_ih_j}$ is state-independent:
\begin{equation}
D^{(2)}_{h_i h_j}(\mathbf{X}) = \mathbb{E}[P^{h_i h_j}_3].
\end{equation}

\bibliography{references}

\end{document}